\documentclass[pdflatex,sn-mathphys]{sn-jnl}% Math and Physical Sciences Reference Style
\jyear{2023}%

\theoremstyle{thmstyleone}%
\theoremstyle{thmstyletwo}%

\theoremstyle{thmstylethree}%

\usepackage{subcaption}
\usepackage{upgreek}

\algrenewcommand\algorithmicrequire{\textbf{Input:}}
\algrenewcommand\algorithmicensure{\textbf{Output:}}

\begin{document}

\title[Functional Diffusion Maps]{Functional Diffusion Maps}


\author[1]{\fnm{M.} \sur{Barroso}}\email{maria.barrosoh@estudiante.uam.es}

\author[1]{\fnm{C.M.} \sur{Alaíz}}\email{carlos.alaiz@uam.es}
\equalcont{These authors contributed equally to this work.}

\author[2]{\fnm{J.L.} \sur{Torrecilla}}\email{joseluis.torrecilla@uam.es}
\equalcont{These authors contributed equally to this work.}

\author*[1]{\fnm{A.} \sur{Fernández}}\email{a.fernandez@uam.es}
\equalcont{These authors contributed equally to this work.}

\affil*[1]{\orgdiv{Departamento de Ingenier\'{i}a Inform\'{a}tica}, \orgname{Universidad Aut\'{o}noma de Madrid}, \orgaddress{\country{Spain}}}

\affil[2]{\orgdiv{Departamento de Matem\'{a}ticas}, \orgname{Universidad Aut\'{o}noma de Madrid}, \orgaddress{\country{Spain}}}


\abstract{
Nowadays many real-world datasets can be considered as functional, in the sense that the processes which generate them are continuous.
A fundamental property of this type of data is that in theory they belong to an infinite-dimensional space.
Although in practice we usually  receive finite observations, they are still high-dimensional and hence dimensionality reduction methods are crucial.
In this vein, the main state-of-the-art method for functional data analysis is Functional PCA.
Nevertheless, this classic technique assumes that the data lie in a linear manifold, and hence it could have problems when this hypothesis is not fulfilled.
In this research, attention has been placed on a non-linear manifold learning method: Diffusion Maps.
The article explains how to extend this multivariate method to functional data and compares its behavior against Functional PCA over different simulated and real examples.}

\keywords{Unsupervised functional data analysis, Dimensionality reduction methods, Functional diffusion maps, Manifold learning}



\maketitle

\section{Introduction}

In many real-world applications, data must be considered as discretized functions rather than standard vectors, since the process that generates them is assumed to be continuous.
Natural examples are found in medicine, economics, speech recognition or meteorological problems.
This type of data, commonly called functional data, is usually represented by high-dimensional vectors whose coordinates are highly correlated, and thus multivariate methods lead to ill-conditioned problems.
As a consequence, most traditional data analysis tools, such as regression, classification or clustering, are being adapted to functional inputs, as well as some dimensionality reduction techniques.

Functional Data Analysis (FDA;~\cite{VIEU2018134, CUEVAS20141}) is an active field of statistics in which the data of interest are often considered as realizations of stochastic processes over a compact domain. Traditionally, FDA focused on linear dimensionality reduction methods, with special attention given to Functional Principal Component Analysis (FPCA; \cite{Ramsay_FDA}) as an unsupervised technique that captures most of the variance of the original data in a lower-dimensional space.
In particular, in this method the infinite-dimensional random functions are projected onto the lower dimensional subspace generated by the eigenfunctions of the covariance operator.
Moreover, when functional data are expressed as a linear combination of known basis functions, the eigenequation problem can be reduced to a discrete form expressing the eigenfunctions in the same known basis~\cite{representation_theorem}.

Other examples of popular linear algorithms that have been transferred to FDA include Independent Component Analysis (ICA; \cite{VIRTA2020104568}), which attempts to project data onto independent components, and Partial Least Squares (FPLS; \cite{Delaigle2012MethodologyAT}), which aims to identify the underlying latent variables in both the response and predictor variables.

A extreme case of linear projections are variable selection methods that reduce trajectories to their values at a few impact points. In recent years, a multitude of proposals in functional variable selection have emerged, highlighting methods based on the regularization of linear regression models such as Lasso or Dantzig \cite{ANEIROS2022104871, VariableselectionL1regularization}. Other proposals are based on logistic regression \cite{10.1214/10-AOS791}, wrapper methods \cite{Componentwiseclustering}, or the selection of local maxima of dependence measures \cite{Luis_2016}. While all of these proposals have notable strengths, they are inadequate for addressing non-linear problems.

Multivariate statistics offers several non-linear techniques for dimensionality reduction based on the assumption that high dimensional data observed in a $D$-dimensional space $\mathbb{R}^D$ actually lie on (or close to) a $d$-dimensional manifold $\mathcal{M}\subset \mathbb{R}^D$, with $d < D$~\cite{Algorithms_for_manifold_learning}. These techniques can be used for discovering high-density low-dimensional surfaces in unsupervised problems, or as a preprocessing step when applying supervised models, being very useful, for example, for data analysis and visualization~\cite{bengio2006}.
Their goal is to achieve a new representation that captures the structure of the data in a few dimensions while preserving the original local information.
For doing this, most of these methods rely on the spectral analysis of a similarity matrix of a graph previously constructed over the original data.
Another important characteristic of these methods is that they may arrive at a new space where the Euclidean distance between embedded points somehow corresponds to the original information preserved.

Isometric feature mapping (Isomap;~\cite{A_Global_Geometric_Framework_for_nonlinear_Dimensionality_Reduction}), t-distributed Stochastic Neighborhood Embedding (t-SNE;~\cite{Multidimensional_Scaling}), Locally Linear Embedding (LLE;~\cite{LLE}) or Diffusion Maps (DM;~\cite{Diffusion_Maps, An_Introduction_to_Diffusion_Maps}) are some of the main manifold methods used for dimensionality reduction over multivariate data.
They all take advantage of the local linearity of manifolds and create a mapping that preserves local neighborhoods at each point of the underlying manifold.

Out of all of these techniques, only Isomap has been extended to the functional setting \cite{Nonlinear_manifold_representations_for_functional_data}. 
Isomap is an extension of the classical multidimensional scaling metric and it is characterized by the use of geodesic distances induced by a neighborhood graph for estimating the intrinsic geometry of a data manifold.
Despite the fact that Isomap is often easy to visualize and interpret, making it useful for exploratory data analysis and data visualization, the computation of geodesic distances can be computationally expensive. Therefore, there is a need to extend other non-linear dimension reduction methods to the functional context that deal with these difficulties. A promising approach is the utilization of Diffusion Maps, which has the ability to handle more complex data with multiple scales and is much more robust against noise and outliers.

% In recent years, non-linear functional models have been widely studied, for instance, Functional Support Vector Machines (FSVM;~\cite{FSVM}) appeared as a possible non-linear method for data classification, Functional Neural Networks were proposed for regression problems~\cite{FunctionalNeuralNetworks} and Functional Isomap~\cite{Nonlinear_manifold_representations_for_functional_data} emerged as a non-linear dimensionality reduction method based on manifold learning, where an underlying manifold assumption is made over functional data.

% When data are functions, they live in infinite-dimensional spaces, and hence finding low-dimensional representations becomes essential.
% Finding reliable representations of low-dimensional data, specifically in two or three dimensions, is beneficial in real-world problems for visualization, description and general exploration purposes~\cite{Unsupervised_Functional_Data_Analysis}.
% Moreover, these representations can be employed with conventional multivariate machine learning algorithms~\cite{Benchmarking_time_series_classification}.

The current paper specifically focuses on extending this manifold learning technique to the FDA context by proposing a new method: Funcional Diffusion Maps (FDM).
As DM, which has been successfully applied in various multivariate \cite{dm_meteo, dm_audio_visual} and even time-series data \cite{dm_time_coupled, dm_time_series}, FDM has the potential to deliver competitive results in functional data applications, particularly in cases where the data is distributed on specific manifolds.
The main contribution of this research can thus be summarized in extending DM to the functional domain, offering a formalization of DM for functions and comparing FDM with DM, FPCA and Isomap in simulated and real examples. 

The rest of the paper is organized as follows:
Section~\ref{seq:background} presents some ideas and definitions from FDA that will be useful for the subsequent discussion in this article, paying special attention to functional manifolds, which will be the main assumption in DM.
Section~\ref{sec:fdm} formalizes the FDM approach and Section~\ref{sec:examples} illustrates and compares the performance of DM, FPCA, Isomap and FDM on synthetic and real datasets.
Finally, Section~\ref{sec:conclusion} presents some conclusions of this work, as well as possible research lines to further improve it.

\section{Functional Data Analysis}\label{seq:background}

In this work, we consider the typical case in Functional Data Analysis ~\cite{Ramsay_FDA, ferraty_FDA} where we have sample of functions $x_1(t), \dotsc, x_N(t)$, where $t\in \mathcal{J}$, $\mathcal{J}$ is a compact interval, and each $x_i(t)$ is the observation of an independent functional variable $X_i$ identically distributed as $X$. 
It is usual to assume that the functional variable $X$ is a second order stochastic process, $\mathrm{E}[X^2]<\infty$, and takes values in the Hilbert space of square integrable functions $L^2([a,b])$ defined on the closed interval $[a,b] \subset \mathbb{R}$.
Square integrable functions form a vector space and we can define the inner product of two functions by $\langle x,y \rangle = \int_{a}^b x(t) y(t) dt$.
The inner product allows to introduce the notion of distance between functions by the $L^2$-norm $\|x\|^2 = \langle  x,x \rangle = \int_{a}^b x^2(t)dt.$
Therefore, in $L^2$ space, the distance between two functions can be calculated as the norm of their difference, which is expressed as $\|x-y\|$.

\subsection{Functional data representation}

Although each function $x_i$ consists of infinitely many values $x_i(t)$ with $t \in [a,b]$, in practice we only observe a finite set of function values in a set of arguments that are not necessarily equidistant or equal for all functions~\cite{Florence_FPCA}.
A main problem representing functional data is that we can not directly retrieve the underlying function and we need to work with the functions using specific approaches: working directly with discrete data or transforming it into a function using a basis function expansion.
We will briefly sketch below both approaches (for a throughout explanation the reader can refer to~\cite{Ramsay_FDA}).

The first approach consists in using directly the observed values of the functions and it does not require any additional adjustment or assumption.
Let $\{x_i(t)\}_{i=1}^N$ be $N$ underlying functions from which we have obtained our data $\mathrm{x}_i = ( x_i(t_1), \dotsc,x_i(t_M))^{\top}$; we denote our set of data as the $N \times M$ matrix defined as 
\begin{equation}\label{eq:discrete}
    \mathrm{X} = \begin{pmatrix}
        \mathrm{x}_1^{\top}\\
        \mathrm{x}_2^{\top}\\
        \vdots\\
        \mathrm{x}_N^{\top}
    \end{pmatrix} = \begin{pmatrix}
        x_1(t_1) & x_1(t_2) & \cdots & x_1(t_M) \\
        x_2(t_1) & x_2(t_2) & \cdots & x_2(t_M) \\
        \vdots & \vdots & \ddots & \vdots \\
        x_N(t_1) & x_N(t_2) & \cdots & x_N(t_M) \\
    \end{pmatrix}.
\end{equation}

The other possible representation is based on a basis function expansion and consist in representing each function $x_i(t)$ using a basis in $L^2$, such that
\begin{equation*}
x_i(t) = \sum_{k=1}^\infty c_{ik} \phi_k(t),
\end{equation*}
where $\phi_k(t)$ are the basis functions and $c_{ik}$ are the coefficients that determine the weight of each basis function in the representation of $x_i(t)$.

One way to approximate the function is by truncating the basis to the first $K$ elements. This results in smoothing and dimensionality reduction, such that,
\begin{equation*}
    x_i(t) \approx \sum_{k=1}^K c_{ik}\phi_k(t) = \mathrm{c}_i^{\top} \upphi(t),
\end{equation*}
where $\upphi(t) = (\phi_1(t), \dotsc, \phi_K(t))^{\top}$ is the vector containing the functions of the truncated basis and $\mathrm{c}_i = (c_{i1}, \dotsc, c_{iK})^{\top}$ are the coefficients in the new basis.
In this way, the set of $N$ functional data evaluated at time $t$ can be expressed as follows:
\begin{align}
    \mathrm{X}(t) & = \mathrm{C} \mathrm{\upphi}(t) = \begin{pmatrix}
        \mathrm{c}_1^{\top} \\
        \mathrm{c}_2^{\top} \\
        \vdots \\
        \mathrm{c}_N^{\top}
    \end{pmatrix} \begin{pmatrix}
        \phi_1(t) \\
        \phi_2(t) \\
        \vdots \\
        \phi_K(t)
    \end{pmatrix}
    = \begin{pmatrix}
        c_{11} & c_{12} & \cdots & c_{1K} \\
        c_{21} & c_{22} & \cdots & c_{2K} \\
        \vdots & \vdots & \ddots & \vdots \\
        c_{N1} & c_{N2} & \cdots & c_{NK}
    \end{pmatrix} \begin{pmatrix}
        \phi_1(t) \\
        \phi_2(t) \\
        \vdots \\
        \phi_K(t)
    \end{pmatrix},
\end{align}
\label{eq:basis}
where $\mathrm{C}$ is an $N\times K$ matrix.

There are different possible basis, such as Fourier basis, wavelet, B-Spline, polynomial basis, etc.
Depending on the data, the choice of basis can be key to capture as much information as possible about the underlying function.
For example, if the data are periodic, a Fourier basis will be of interest, while if they are not, the B-Spline basis is a common choice.

\subsection{Functional manifolds}

A functional manifold $\mathcal{M}$ is a differential manifold which is locally diffeomorphic to $\mathbb{R}^d$~\cite{differential-manifold}.
These manifolds are expressed in terms of an atlas consisting of a set of charts $\{(U_\alpha, \varphi_\alpha)\}$, where $U_\alpha$ are open sets covering $\mathcal{M}$, and $\varphi_\alpha: U_\alpha \to \mathbb{R}^d$, the coordinate maps, are diffeomorphisms from $U_\alpha$ to open sets of $\mathbb{R}^d$.
These coordinate maps  are commonly used to describe the geometry of the manifold.

In the context of FDA, non-linear dimensionality reduction methods uses ``simple'' functional manifolds~\cite{Nonlinear_manifold_representations_for_functional_data}, which are differential manifolds that are locally isometric to $\mathbb{R}^d$.
These manifolds are characterized by a single chart $(U, \varphi)$, where $U$ is an open set covering $\mathcal{M}$, and the coordinate map $\varphi: U\to\mathbb{R}^d$ is an isometry from $U$ to an open set of $\mathbb{R}^d$.
A typical example of a ``simple" functional manifold is the well-known Swiss roll manifold (see Figure \ref{fig:toy-datasets}).

In our research, we will focus on functions in these ``simple'' functional manifolds, specifically those that belong to $\mathcal{M}\subset L^2([a,b])$, where $[a,b] \subset \mathbb{R}$ is a closed interval.
We will use manifold learning methods to approximate $\varphi$ and understand the inherent structure of the functional data.

\section{Functional Diffusion Maps}\label{sec:fdm}

In this work, we propose a new Functional Manifold Learning technique, which we call Functional Diffusion Maps, that aims to find low-dimensional representations of $L^2$ functions on non-linear functional manifolds. 
It first builds a weighted graph based on pairwise similarities between functional data and then, just like Diffusion Maps does, uses a diffusion process on the normalised graph to reveal the overall geometric structure of the data at different scales. 

% Even though  giving a geometric interpretation of similarity between functional data is sometimes not possible, FDM retains the advantages of multivariate DM as it is robust to noise perturbation. 
% FDM is also a very flexible algorithm that allows fine-tuning of parameters that influence the resulting embedding.

In the following, the mathematical framework for DM is detailed, and then a generalization for functional data is proposed.

\subsection{Mathematical framework for Diffusion Maps} \label{sec:diffusion-maps}

Diffusion Maps is a non-linear dimensionality reduction algorithm introduced by Coifman and Lafon~\cite{Diffusion_Maps} which focus on discovering the underlying manifold from which multivariate data have been sample.
DM computes a family of embeddings of a dataset into a low-dimensional Euclidean subspace whose coordinates are computed from the eigenvectors and eigenvalues of a diffusion operator on the data.

In particular, let $\mathcal{X}=\{\mathrm{x_1}, \dotsc,\mathrm{x_N}\}$ be our multivariate dataset, such that $\mathcal{X}$ is lying along a manifold $\mathcal{M} \subset \mathbb{R}^D$.
A random walker will be defined on the data, so the connectivity or similitude between two data points $\mathrm{x}$ and $\mathrm{y}$ can be used as the probability of walking from $\mathrm{x}$ to $\mathrm{y}$ in one step of the random walk.
Usually, this probability is specified in terms of a kernel function of the two points, $k:\mathbb{R}^D\times \mathbb{R}^D\to \mathbb{R}$, which has the following properties:
\begin{itemize}
    \item $k$ is symmetric: $k(\mathrm{x},\mathrm{y}) = k (\mathrm{y},\mathrm{x})$;
    \item $k$ is positivity preserving: $k(\mathrm{x}, \mathrm{y})\geq 0$ $\forall \mathrm{x},\mathrm{y}\in\mathcal{X}$.
\end{itemize}

Let $\mathrm{G} = (\mathcal{X} , \mathrm{K})$ be a weighted graph where $\mathrm{K}$ is an $N\times N$ kernel matrix whose components are $k_{ij} = k(\mathrm{x}_i,\mathrm{x}_j)$. 
This graph is connected and captures some geometric features of interest in the dataset. Now, to build a random walk, $\mathrm{G}$ has to be normalized.

Since the graph depends on both the geometry of the underlying manifold and the data distribution on the manifold, in order to control the influence of the latter, DM uses a density parameter $\alpha\in [0,1]$ in the normalization step. 
The normalized graph $\mathrm{G}'=(\mathcal{X} , \mathrm{K}^{(\alpha)})$ uses the weight matrix $\mathrm{K}^{(\alpha)}$ obtained by normalizing $\mathrm{K}$:
\begin{equation*}
    k_{ij}^{(\alpha)} = \frac{k_{ij}}{d_i^\alpha d_j^\alpha},
\end{equation*}
where $d_i = \sum_{j=1}^N k_{ij}$ is the degree of the graph and the power $d_i^\alpha$ approximates the density of each vertex. 

Now, we can create a Markov chain on the normalized graph whose transition matrix $\mathrm{P}$ is defined by
\begin{equation*}
    p_{ij} = \frac{k_{ij}^{(\alpha)}}{d_i^{(\alpha)}},
\end{equation*}
where $d_i^{(\alpha)}$ is the degree of the normalized graph,
\begin{equation*}
    d_i^{(\alpha)} = \sum_j k_{ij}^{(\alpha)} = \sum_j \frac{k_{ij}}{d_i^\alpha d_j^\alpha}.
\end{equation*}

Each component $p_{ij}$ of the transition matrix $\mathrm{P}$ provides the probabilities of arriving from node $i$ to node $j$ in a single step. 
By taking powers of the $\mathrm{P}$ matrix we can increase the number of steps taken in the random walk. 
Thus, the components of $P^T$, namely $p_{ij}^T$, represent the sum of the probabilities of all paths of length $T$ from $i$ to $j$. 
This defines a \textit{diffusion process} that reveals the global geometric structure of $\mathcal{X}$ at different scales.

The transition matrix of a Markov chain has a stationary distribution, given by
\begin{equation*}
    \pi_i = \frac{d_i^{(\alpha)}}{\sum_k d_k^{(\alpha)}},
\end{equation*}
that satisfies the stationary property as follows:
\begin{equation*}
    \sum_{i=1}^N \pi_i p_{ij} = \sum_{i=1}^N \frac{d_i^{(\alpha)}}{\sum_k d_k^{(\alpha)}} \frac{k_{ij}^{(\alpha)}}{d_i^{(\alpha)}}
    = \sum_{i=1}^N \frac{k_{ij}^{(\alpha)}}{\sum_k d_k^{(\alpha)}} = \frac{d_j^{(\alpha)}}{\sum_k d_k^{(\alpha)}} = \pi_j.
\end{equation*}

Since the graph is connected, the associated stationary distribution is unique. 
Moreover, since $\mathcal{X}$ is finite, the chain is ergodic. 
Another property of the Markov chain is that it is reversible, satisfying 
\begin{equation*}
    \pi_i p_{ij} = \frac{d_i^{(\alpha)} }{\sum_k d_k^{(\alpha)}} \frac{k_{ij}^{(\alpha)}}{d_i^{(\alpha)}}  = \frac{k_{ij}^{(\alpha)}}{\sum_k d_k^{(\alpha)}} = \frac{k_{ji}^{(\alpha)}}{\sum_k d_k^{(\alpha)}} = \frac{d_j^{(\alpha)} }{\sum_k d_k^{(\alpha)}} \frac{k_{ji}^{(\alpha)}}{d_i^{(\alpha)}} = \pi_j p_{ji}.
\end{equation*}

Taking into account all these properties, we can now define a diffusion distance $\mathrm{D}_T$ based on the geometrical structure of the obtained diffusion process. 
The metric measures the similarities between data as the connectivity or probability of transition between them, such that
\begin{equation}\label{eq:diffusion_distance}
    \mathrm{D}_T^2 (\mathrm{x}_i,\mathrm{x}_j) = \|p_{i\cdot}^T - p_{j\cdot}^T\|^2_{\mathrm{L}^2(\frac{1}{\pi})} = \sum_k \frac{\left(p_{ik}^T - p_{jk}^T\right)^2}{\pi_k},
\end{equation}
where $\|\frac{1}{\pi}\|_{L^2}$  represents the euclidean norm weighted by the stationary distribution $\pi$, which takes into account the local data point density. This metric is also robust to noise perturbation, since it is defined as an average of all the paths of length $T$. Therefore, $T$ plays the role of a scale parameter and $\mathrm{D}_T^2 (\mathrm{x}_i,\mathrm{x}_j)$ will be small if there exist a lot of paths of length $T$ that connect $\mathrm{x}_i$ and $\mathrm{x}_j$.

\begin{figure}[ht]
\centering
    \begin{subfigure}[b]{0.49\textwidth}
        \centering
        \includegraphics[width=\textwidth,height=1.7in]{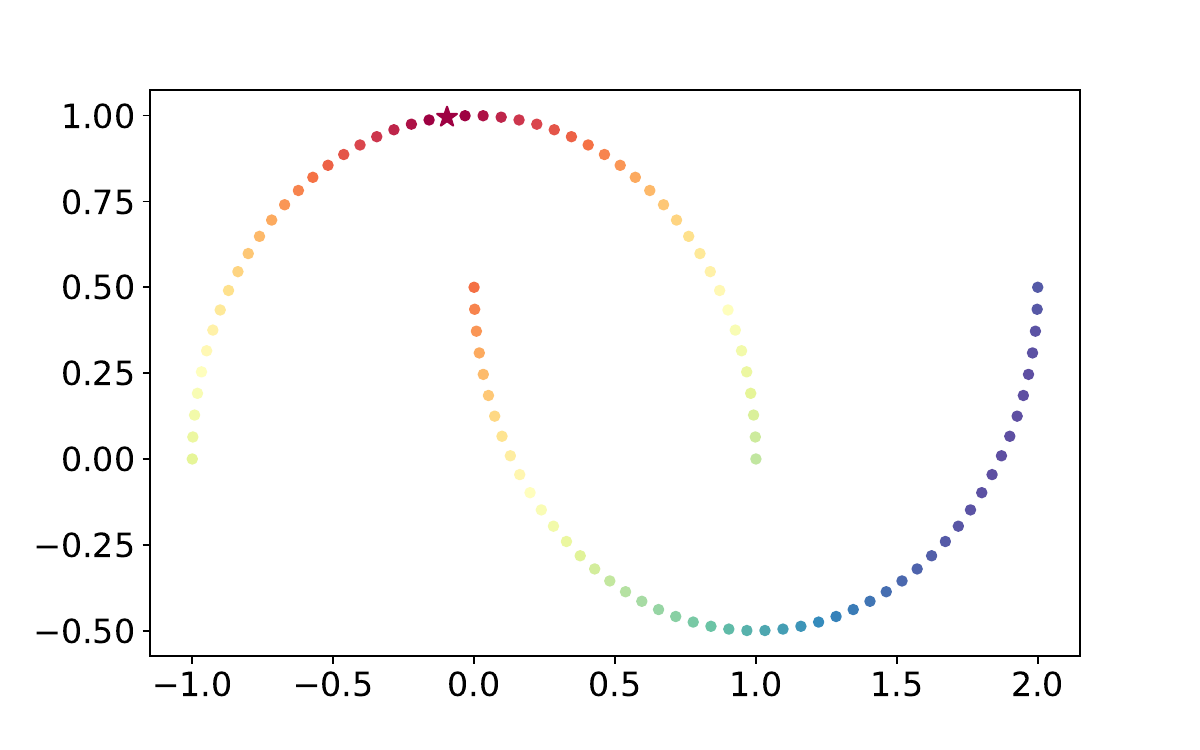}
        \caption{$L^2$ distances.}
    \end{subfigure}%
    ~ 
    \begin{subfigure}[b]{0.49\textwidth}
        \centering
        \includegraphics[width=\textwidth,height=1.7in]{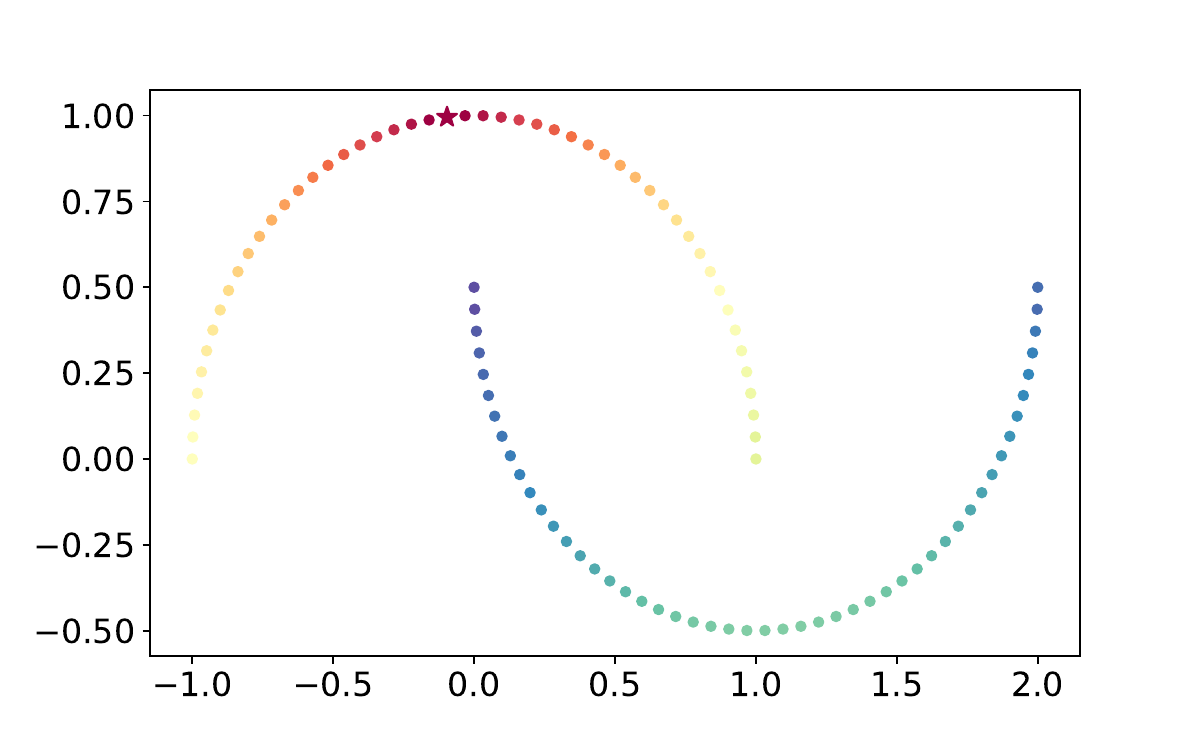}
        \caption{Diffusion distances.}
    \end{subfigure}
    \caption{A comparison between the $L^2$ distance and the diffusion distance over the Moons dataset.}
    \label{fig:moon-distances}
\end{figure}

Figure~\ref{fig:moon-distances} shows a visual example of the diffusion distances and the $L^2$ distances over the classic Moons dataset, which consists of two half circles interspersed. 
A spectral color bar has been used, with warm colors indicating nearness between the data and cold colors representing remoteness.

Clearly, Moons data live in a non-linear manifold. 
In the picture, the distances between the point marked with a star and the rest are displayed.
It can be seen that the diffusion distance approximates the distance in the manifold, so that points on the same moon are nearer among them than to points on the other moon.
In contrast, $L^2$ distances are independent of the moons and are not convenient when data live in a non-linear manifold such as this one.

Spectral analysis of the Markov chain  allows us to consider an alternative formulation of the diffusion distance that uses eigenvalues and eigenvectors of $\mathrm{P}$.

As detailed in~\cite{Diffusion_Maps}, even though $\mathrm{P}$ is not symmetric, it makes sense to perform its spectral decomposition using its left and right eigenvectors, such that
\begin{equation*}
    p_{ij}=\sum_{l\geq 0} \lambda_l (\psi_l)_i (\varphi_l)_j.
\end{equation*}
The eigenvalue $\lambda_0=1$ is discarded for $P$ since $\psi_0$ is a vector with all its values equal to one.
The other eigenvalues $\lambda_1, \lambda_2, \dotsc,\lambda_{N-1}$ tend to $0$ and satisfy $\lambda_l<1$ for all $l \geq 1$.
Now, the diffusion distance can be rewritten using the new representation of $\mathrm{P}$:
\begin{equation*}
\begin{split}
    \mathrm{D}_T^2 (\mathrm{x}_i,\mathrm{x}_j)  &= \sum_k \frac{1}{\pi_k} \left(\sum_{l\geq1}\lambda_l^T (\psi_l)_i (\varphi_l)_k - \sum_{l\geq1}\lambda_l^T (\psi_l)_j (\varphi_l)_k\right)^2 \\
    &=\sum_{l\geq1}\lambda_l^{2T}\left((\psi_l)_i - (\psi_l)_j\right)^2  \sum_k \frac{(\varphi_l)_k^2}{\pi_k} \\
    &= \sum_{l\geq1}\lambda_l^{2T} \left((\psi_l)_i - (\psi_l)_j\right)^2  \sum_k \frac{((\phi_l)_k \sqrt{\pi_k})^2}{\pi_k} \\
    &= \sum_{l\geq1}\lambda_l^{2T} \left((\psi_l)_i - (\psi_l)_j\right)^2 .
\end{split}
\end{equation*}
Since $\lambda_l\to 0$ when $l$ grows, the diffusion distance can be approximated using the first $L$ eigenvalues and eigenvectors. 
One possibility is to select $L$ such that
\begin{equation*}
    L = \max \{l:\lambda_l^T>\delta \lambda_1^T\},
\end{equation*}
where $\delta$ is a parameter that controls the precision in the approximation.
Thus, the diffusion distance approximated by $L$ is expressed as
\begin{equation*}
    \mathrm{D}_T^2(\mathrm{x}_i, \mathrm{x}_j) \approx \sum_{l=1}^{L}\lambda_l^{2T} \left((\psi_l)_i - (\psi_l)_j\right)^2.
\end{equation*}
Finally, the diffusion map $\Psi_T: \mathcal{X}\to\mathbb{R}^L$ is defined as
\begin{align}
    \Psi_T(\mathrm{x}_i) = \begin{pmatrix}
           \lambda_1^T \psi_1(\mathrm{x}_i) \\
           \lambda_2^T \psi_2(\mathrm{x}_i) \\
           \vdots \\
           \lambda_{L}^T \psi_{L}(\mathrm{x}_i)
         \end{pmatrix},
\end{align}
With this definition, diffusion maps family $\{\Psi_T\}_{T\in\mathbb{N}}$ projects data points into a Euclidean space $\mathbb{R}^{L}$ so that diffusion distance on the original space can be approximated by the Euclidean distance of $\Psi_T$ projections in $\mathbb{R}^{L}$,
\begin{equation*}
    \mathrm{D}_T^2(\mathrm{x}_i(t), \mathrm{x}_j(t)) \approx \sum_{l=1}^{L}\lambda_l^{2T} \left((\psi_l)_i - (\psi_l)_j\right)^2 = \|\Psi(\mathrm{x}_i) - \Psi(\mathrm{x}_j)\|^2,
\end{equation*}
preserving the local geometry of the original space.

\subsection{Extending Diffusion Maps to functional data}

Let $X$ be a centered square integrable functional variable of the Hilbert space  $ L^2([a,b])$, where $[a,b]$ is a compact interval.
Let $\mathcal{X} = \{x_1(t), \dotsc, x_N(t)\}$ be the observations of $N$  independent functional variables $X_1, \dotsc, X_N$ identically distributed as $X$.
We assume that $\mathcal{X}$ lies on a functional manifold $\mathcal{M}\subset L^2([a,b])$. 

Before defining a diffusion process over $\mathcal{X}$, a random walk over a weighted graph has to be built.
Graph vertices are functions $x_i(t) \in \mathcal{X}$ and weights $k_{ij}$ are given by a symmetric positive-definite Kernel operator $\mathcal{K}:L^2([a,b])\times L^2([a,b])\to \mathbb{R}$, so that the weight between two vertices $x_i(t)$ and $x_j(t)$ is $k_{ij} = \mathcal{K}(x_i(t), x_j(t))$. 
Thus, the weighted graph is $\mathrm{G}=(\mathcal{X}, \mathrm{K})$, where $\mathrm{K}$ is an $N\times N$ kernel matrix resulted from evaluating $\mathcal{K}$ on $\mathcal{X}$. 
The kernel operator defines a local measure of similarity within a certain neighborhood, so that outside the neighborhood the function quickly goes to zero.
The standard kernel used to compute the similarity between functional data is the Radial Basis Function (RBF) kernel, which is defined as:
\begin{equation*}
    \mathcal{K}(x_i(t), x_j(t)) = \exp{\frac{-\|x_i(t)-x_j(t)\|^2_{ L^2}}{2\sigma^2}},
\end{equation*} 
where the size of the local neighborhood is determined by the hyperparameter $\sigma$.
Another classical option is the Laplacian kernel, which is defined as:
\begin{equation*}
    \mathcal{K}(x_i(t), x_j(t)) = \exp{\frac{-\|x_i(t)-x_j(t)\|_{L^1}}{\sigma^2}}.
\end{equation*}
These kernels satisfy the property $\mathcal{K}(x(t),y(t))=\mathcal{\hat{K}}(x(t)-y(t))$, i.e. the kernel only depends on the difference between both elements.

Once the functional data graph $\mathrm{G}$ has been constructed, the algorithm performs the same operations as multivariate DM. 
The full methodology of the method is presented in Algorithm~\ref{alg:fdm}. Furthermore, the Python implementation of the FDM method is available at \url{https://github.com/mariabarrosoh/functional-diffusion-maps/}.

\begin{algorithm}
\caption{FDM}\label{alg:fdm}
\begin{algorithmic}[1]
\Require
\Statex $L$ -- Embedding dimension
\Statex $T$ -- Steps in random walk
\Statex $\alpha$ -- Density parameter
\Statex $\mathcal{K}$ -- Kernel operator $\mathcal{K}:L^2([a,b])\times L^2([a,b])\to \mathbb{R}$
\Statex $\mathcal{X}  = \{x_1(t), \dots, x_N(t)\}$ -- Functional data set
\Ensure
\Statex $\Psi_T(\mathcal{X} )$ -- Embedded functional data \\
Construct the weighted graph $\mathrm{G}=(\mathcal{X} , \mathrm{K})$, where $\mathrm{K}$ is a positive and symmetric matrix with $k_{ij}= \mathcal{K}(x_i(t), x_j(t))$. \\
Compute the density of each vertex: $d_i^{\alpha} = \left(\sum_{j=1}^N k_{ij}\right)^\alpha.$  \\
Construct the normalized graph $\mathrm{G}'=(\mathcal{X} , \mathrm{K}^{(\alpha)})$ with $k_{ij}^{(\alpha)} = \frac{k_{ij}}{d_i^\alpha d_j^\alpha}.$  \\
Define the transition matrix $\mathrm{P}$ with $p_{ij} = \frac{k_{ij}^{(\alpha)}}{d_i^{(\alpha)}}$, where $d_i^{(\alpha)}=\sum_j k_{ij}^{(\alpha)}.$  \\
Obtain the eigenvalues $\{\lambda_l\}_{l\geq0}$ and the right eigenvectors $\{\psi_l\}_{l\geq 0}$ of $\mathrm{P}$ such that 
        \[
          \begin{cases}
            1      &= \lambda_0 > \lambda_1 \geq \dotsc\\
            \mathrm{P}\psi_l &= \lambda_l \psi_l.
          \end{cases}
        \]
        \\
Calculate the diffusion maps  $$\Psi_T(x_i(t)) = \begin{pmatrix}
           \lambda_1^T \psi_1(x_i(t)) \\
           \lambda_2^T \psi_2(x_i(t)) \\
           \vdots \\
           \lambda_{L}^T \psi_{L}(x_i(t))
         \end{pmatrix} \quad \forall x_i(t)\in \mathcal{X}.$$
\end{algorithmic}
\end{algorithm}
\bigskip

\section{Examples and simulation study}\label{sec:examples}

In this section we apply the FDM technique described above to synthetic and classic functional datasets. 
We compare the performance of FDM against its multivariate version and we also evaluate the performance of FDM alongside other FDA techniques, including FPCA and Isomap, to determine its efficacy.

Both DM and FDM require an initial analysis to identify the suitable parameters to cluster the data or reveal the structure of the manifold where the data is supposed to be embedded. 
To achieve this, a grid search is performed in each experiment using the hyperparameters specified in Table~\ref{tab:hyperparameters}.
\begin{table}[ht]
    \centering
    \caption{DM and FDM hyperparameter grid used for finding the best values for the different dataset.}
    \label{tab:hyperparameters}
    \begin{tabular}{c c}
\toprule
\textbf{Hyperparam.} & \textbf{Values} \\
\midrule
$\alpha$ & $\{k/4 : 0 \leq k \leq 4, k\in\mathbb{N}\}$ \\
$\sigma$ & $\{k/10 : 1 \leq k \leq 10, k\in\mathbb{N}\}$ \\
$k$ & $\{\mbox{RBF},\mbox{Laplacian}\}$ \\
\bottomrule
\end{tabular}
\end{table}

The functional version of the Isomap method has been used in practical applications by directly discretizing the values \cite{Unsupervised_Functional_Data_Analysis}. This technique only requires to set the number of neighbors needed to create the graph.
To determine the optimal number of neighbors, we perform a grid search over the set $\{5k : 1 \leq k \leq 5, k\in\mathbb{N}\}$.

\subsection{Cauchy densities data}

The purpose of this first experiment is to demonstrate that the straightforward implementation of Isomap and Diffusion Maps on functional data may not be sufficient if the functions are not equally spaced and the metric employed is unsuitable.

\begin{figure}[ht]
    \centering
    \includegraphics[scale=0.33]{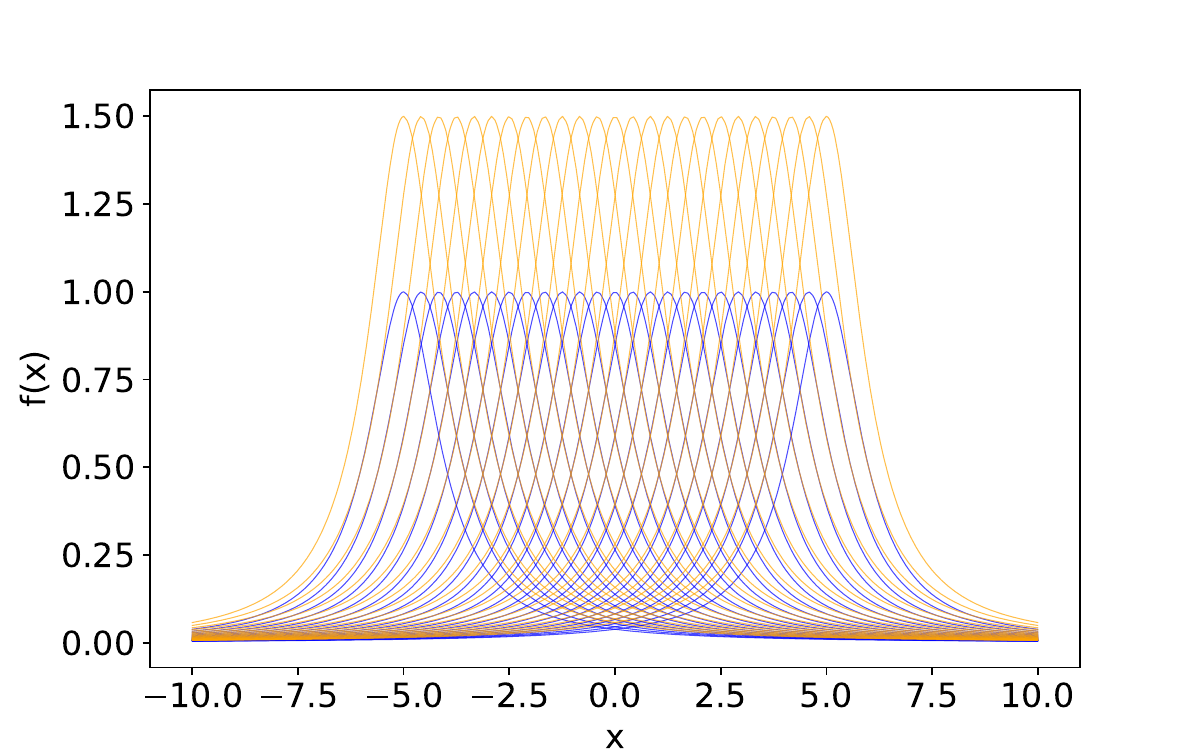}
    \caption{Cauchy densities data.}
    \label{fig:cauchy-fd}
\end{figure}

To illustrate this assertion, we apply Isomap, DM and FDM in a synthetic example consisting of 50 not equally spaced Cauchy probability density functions with $\gamma=1.0$.
In particular, we generate 25 Cauchy densities with amplitude $1.0$ regularly centered in the interval $[-5,5]$ and 25 Cauchy densities with amplitude $1.5$ regularly centered in the same interval. 
The functions are partitioned into equally spaced sections with 100 data points in each of the following intervals: $[-10,-5]$, $(-5,5)$, and $[5,10]$, so that the point density in the middle interval is half that of the other intervals.
The functions generated are displayed in Figure~\ref{fig:cauchy-fd}. 

After evaluating all possible parameter settings for Isomap, DM and FDM on the Cauchy densities, ee found that 15 neighbors are optimal for the Isomap method. A RBF kernel with a scale parameter of $\sigma = 0.6$ and a density parameter of $\alpha = 1.0$ works better for the DM embedding, while a RBF kernel with $\sigma = 0.1$ and $\alpha = 0.0$ is suitable for the FDM technique.

\begin{figure}[ht]
     \centering
     \begin{subfigure}[b]{0.48\textwidth}
         \centering
         \includegraphics[width=\textwidth,height=1.7in]{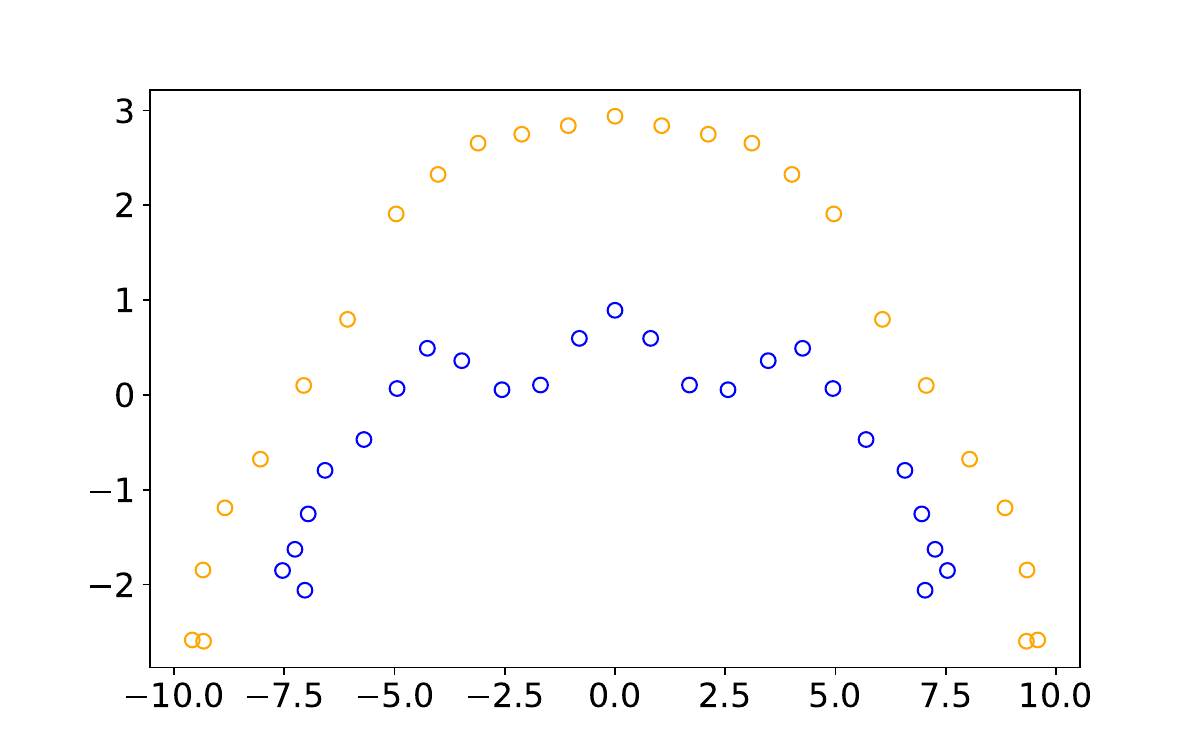}
         \caption{Isomap scores.}
     \end{subfigure}
     \hfill
     \begin{subfigure}[b]{0.48\textwidth}
         \centering
         \includegraphics[width=\textwidth,height=1.7in]{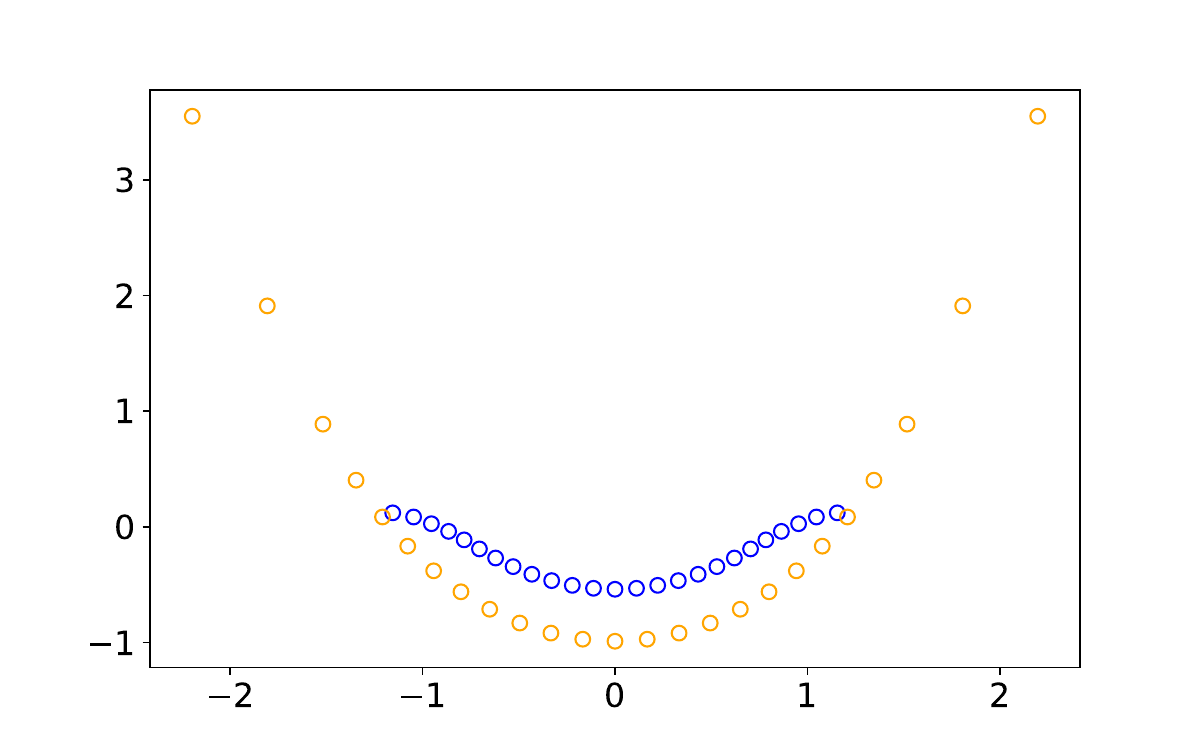}
         \caption{DM scores.}
     \end{subfigure}
     \hfill
     \begin{subfigure}[b]{0.48\textwidth}
         \centering
         \includegraphics[width=\textwidth,height=1.7in]{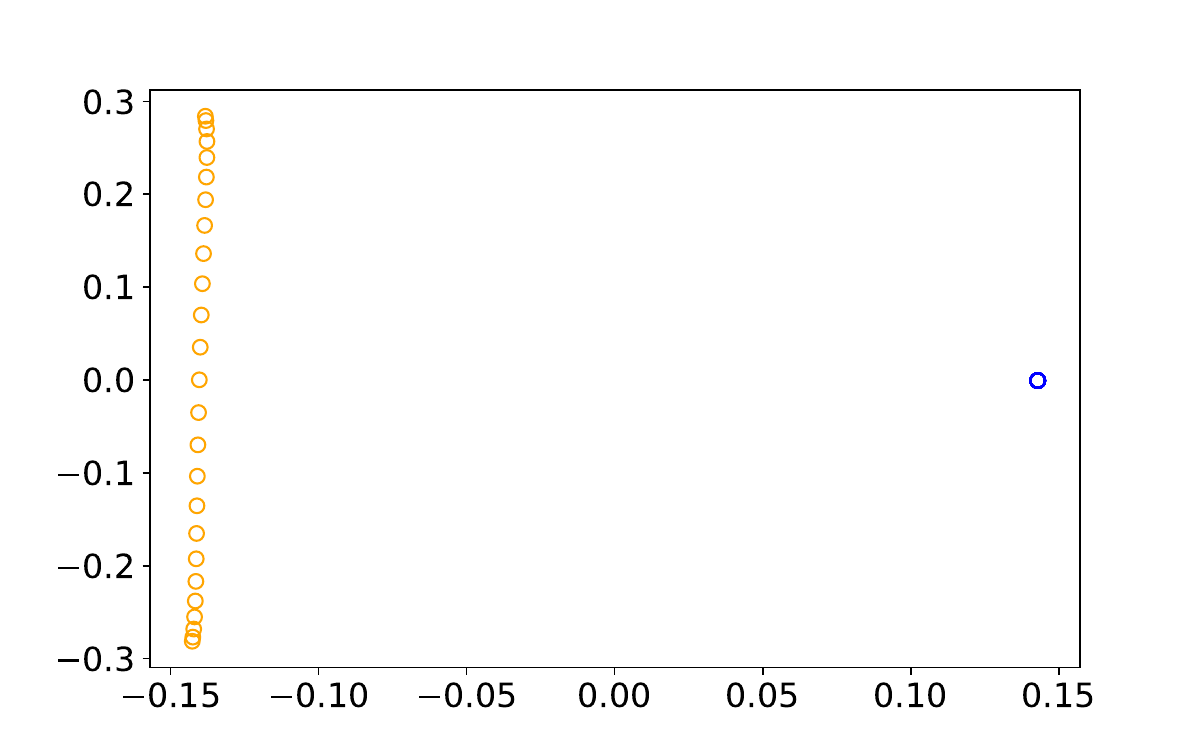}
         \caption{FDM scores.}
     \end{subfigure}
        \caption{Isomap, DM and FDM scores in the first two components for the Cauchy densities dataset.}
    \label{fig:cauchy-embedding}
\end{figure}

Figure~\ref{fig:cauchy-embedding} shows the scatterplots of the multivariate and functional scores in the first two components for the Cauchy densities dataset. 

Both Isomap and DM scores exhibit a semi-circular shape, with the two function classes located in close proximity. 
In contrast, the FDM scores exhibit a complete separation of function classes.
Consequently, we can deduce that the Isomap and multivariate DM may not be an appropriate method for analyzing non-equispaced functions, for which the functional version of DM provides an essential alternative.
The main reason behind this discrepancy lies in the computation of distances.

\subsection{Moons and Swiss Roll data}

After establishing the benefits of using FDM instead of DM for functional data, we aim to contrast FDM with FPCA, wich is the most popular dimensionality reduction method in FDA by far, by evaluating their performance on the functional versions of the Moons and Swiss Roll datasets.

The Moons dataset shown in Subsection~\ref{sec:diffusion-maps} is typically used to visualize clustering algorithms. 
Instead, the Swiss Roll dataset is typically used to test dimensionality reduction techniques. 
Both of them are common examples where non-linearity in the data makes multivariate PCA perform poorly, and therefore, manifold learning techniques are preferable.

\begin{figure}[ht]
     \centering
     \begin{subfigure}[b]{0.45\textwidth}
         \centering
         \includegraphics[width=\textwidth,height=1.7in]{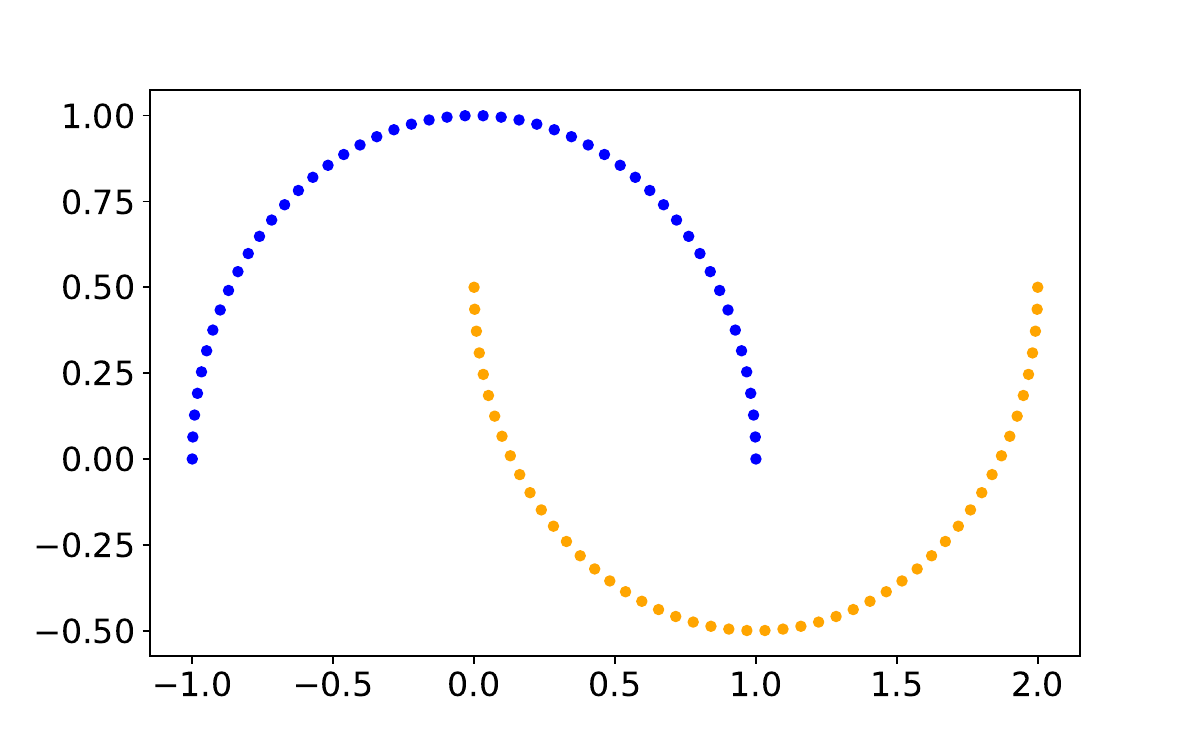}
         \caption{Moons data.}
     \end{subfigure}
     \hfill
     \begin{subfigure}[b]{0.45\textwidth}
         \centering
         \includegraphics[width=\textwidth,height=1.7in]{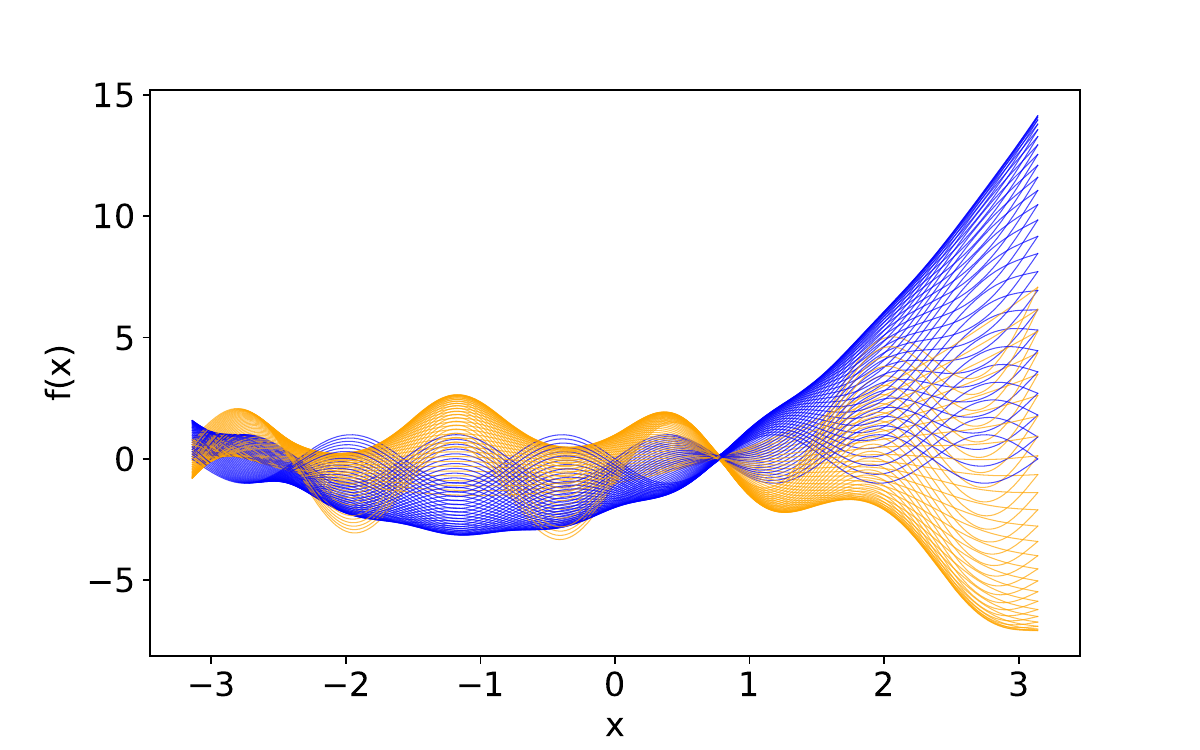}
         \caption{Functional Moons data.}
     \end{subfigure}
     \hfill
     \begin{subfigure}[b]{0.45\textwidth}
         \centering
         \includegraphics[width=\textwidth,height=1.7in]{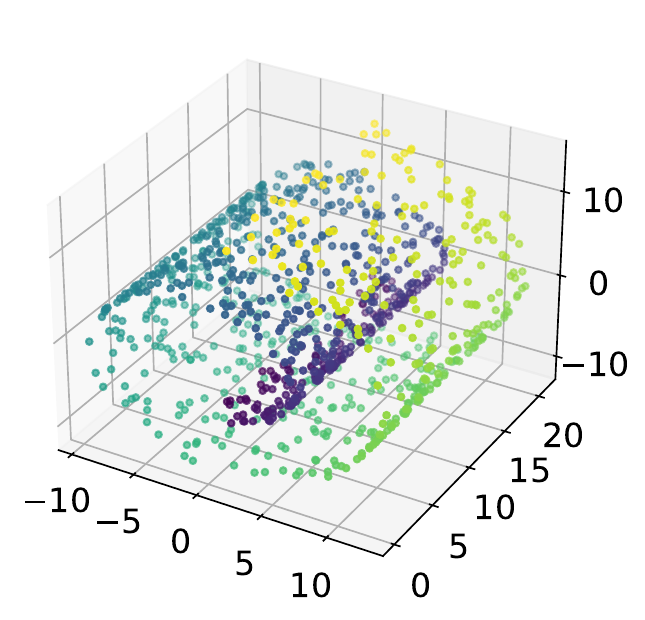}
         \caption{Swiss Roll data.}
     \end{subfigure}
     \hfill
     \begin{subfigure}[b]{0.45\textwidth}
         \centering
         \includegraphics[width=\textwidth,height=1.7in]{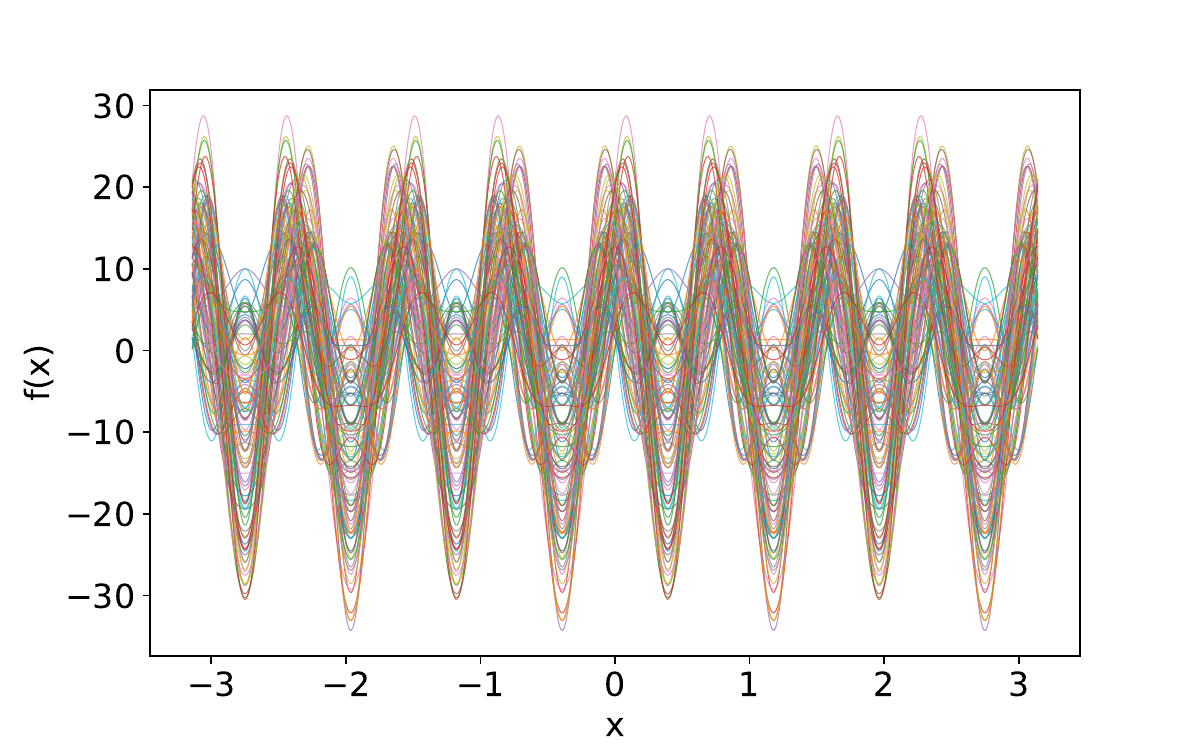}
         \caption{Functional Swiss Roll data.}
     \end{subfigure}

    \caption{Left: Multivariate Moons and Swiss Roll data. Right: Functional version of Moons and Swiss Roll data.}
    \label{fig:toy-datasets}
\end{figure}

Figure~\ref{fig:toy-datasets} shows, at the left, multivariate Moons and Swiss Roll datasets generated without noise. 
To get the functional version of these datasets, which are shown in the right panel of Figure~\ref{fig:toy-datasets}, the features of multivariate data are used as the coefficient of a chosen functional basis. 
Specifically, we represent functional Moons data using the following non-orthogonal basis function:
\[
 \upphi(x) = \{\phi_1(x) = \sin(4x), \phi_2 = x^2 + 2x - 2\} ;
 \]
 for the functional Swiss Roll data we use instead the non-orthogonal basis function:
 \[
  \upphi(x) = \{\phi_1(x) = \sin(4x), \phi_2 = \cos(8x), \phi_3 = \sin(12x)\}.
\]

Since FPCA and FDM act on the basis function coordinates, we expect to obtain the same results as the multivariate ones, except the effect for the inner product between the chosen basis functions.

After evaluating all possible parameter configurations (see Table \ref{tab:hyperparameters}), we found that a RBF kernel with a scale parameter of $\sigma = 0.2$ and a density parameter of $\alpha = 0.5$ works better for the Moons dataset, while a RBF kernel with $\sigma=0.6$ and $\alpha=1.0$ is more suitable for the Swiss Roll dataset.

\begin{figure}
     \centering
     \begin{subfigure}[b]{0.45\textwidth}
         \centering
         \includegraphics[width=\textwidth,height=1.7in]{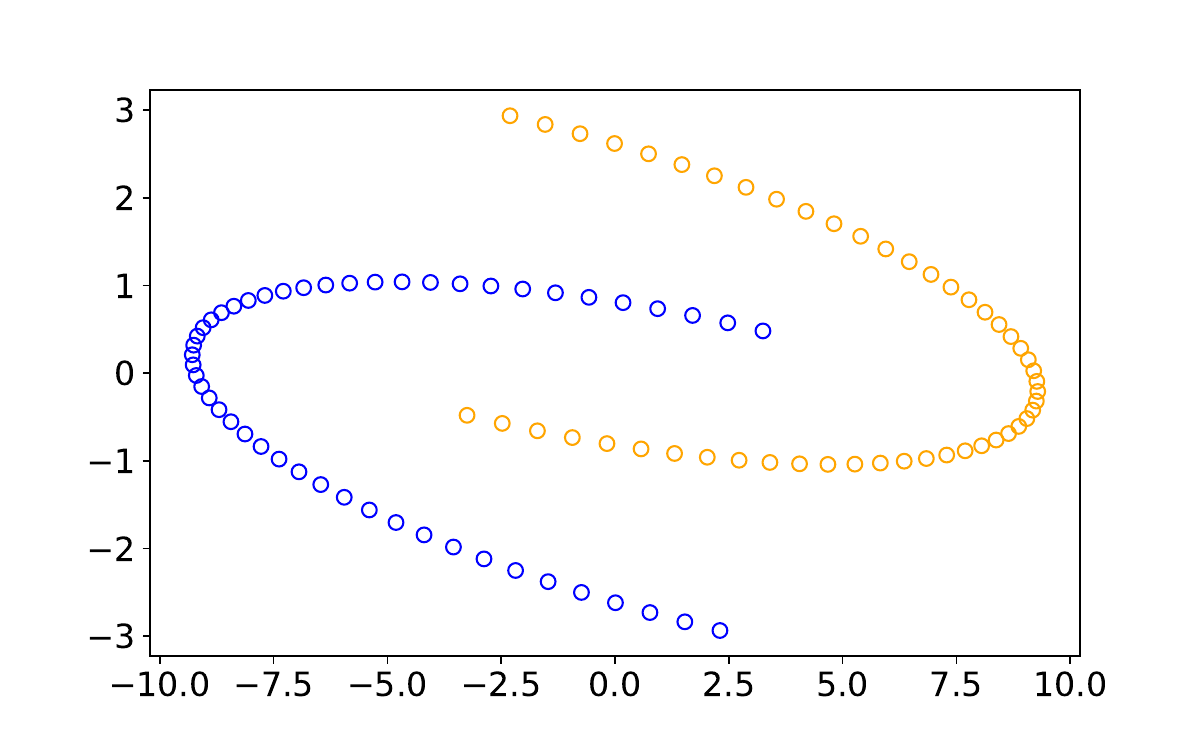}
         \caption{FPCA scores.}
     \end{subfigure}
     \hfill
     \begin{subfigure}[b]{0.45\textwidth}
         \centering
         \includegraphics[width=\textwidth,height=1.7in]{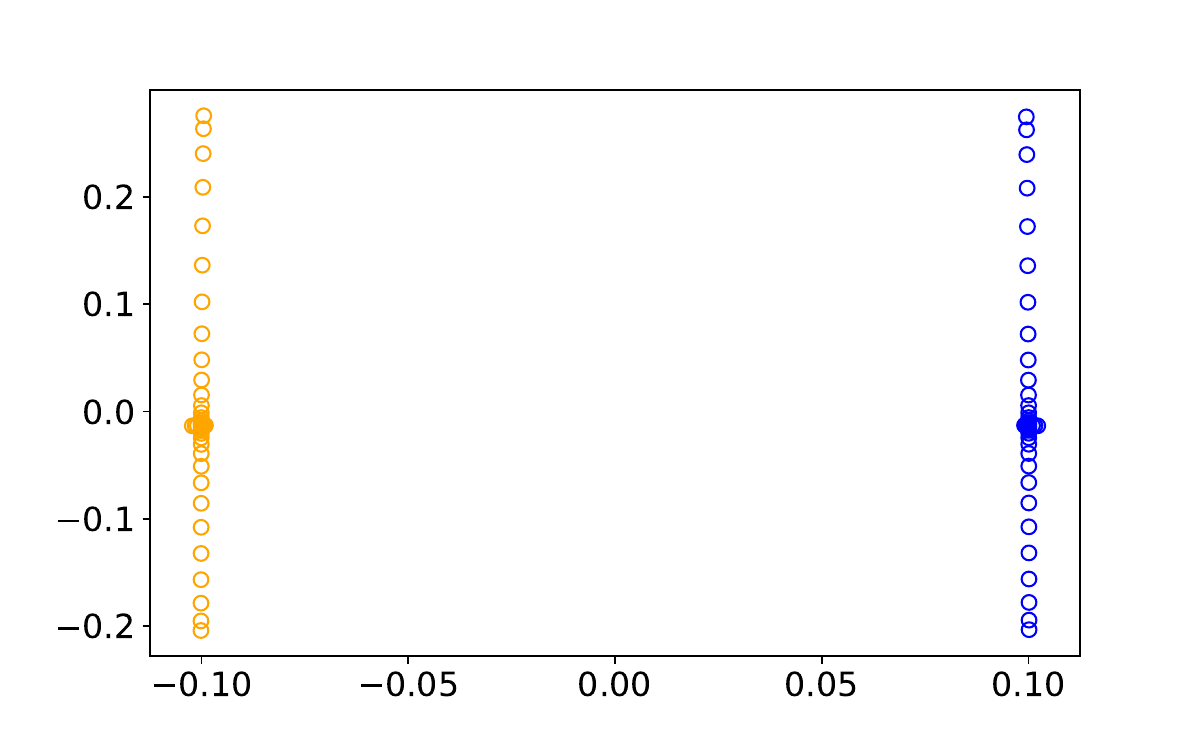}
         \caption{FDM scores.}
     \end{subfigure}
     \hfill
        \begin{subfigure}[b]{0.45\textwidth}
         \centering
         \includegraphics[width=\textwidth,height=1.7in]{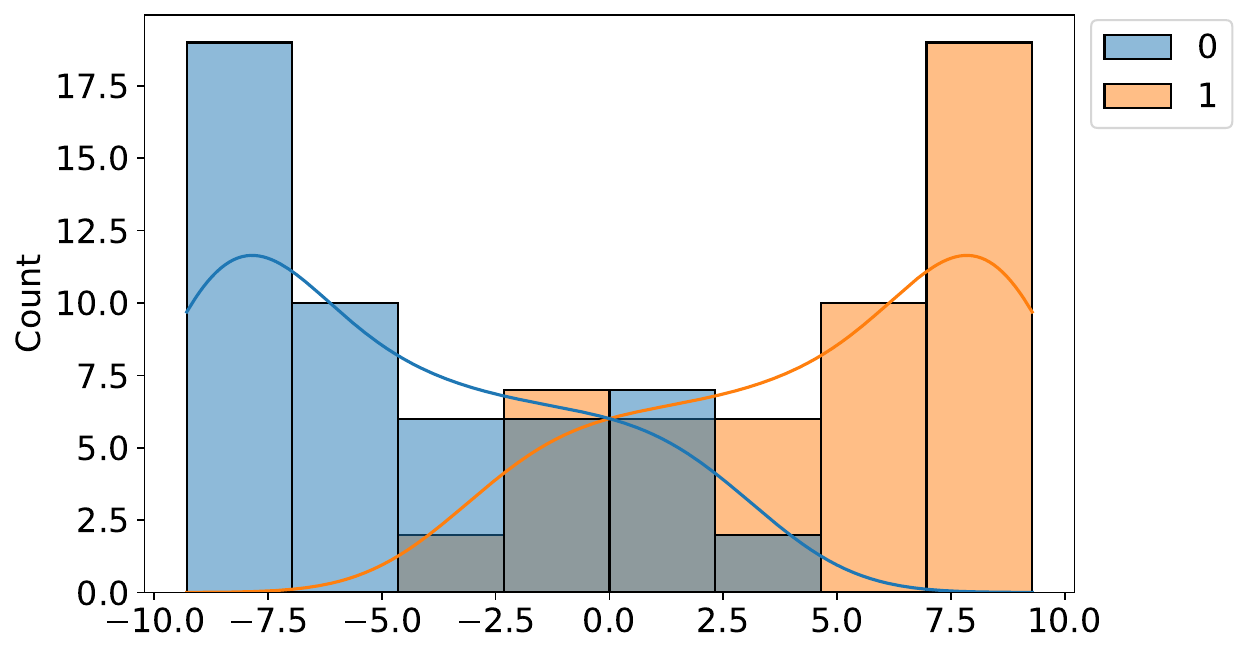}
         \caption{FPCA scores histogram.}
     \end{subfigure}
     \hfill
     \begin{subfigure}[b]{0.45\textwidth}
         \centering
         \includegraphics[width=\textwidth,height=1.7in]{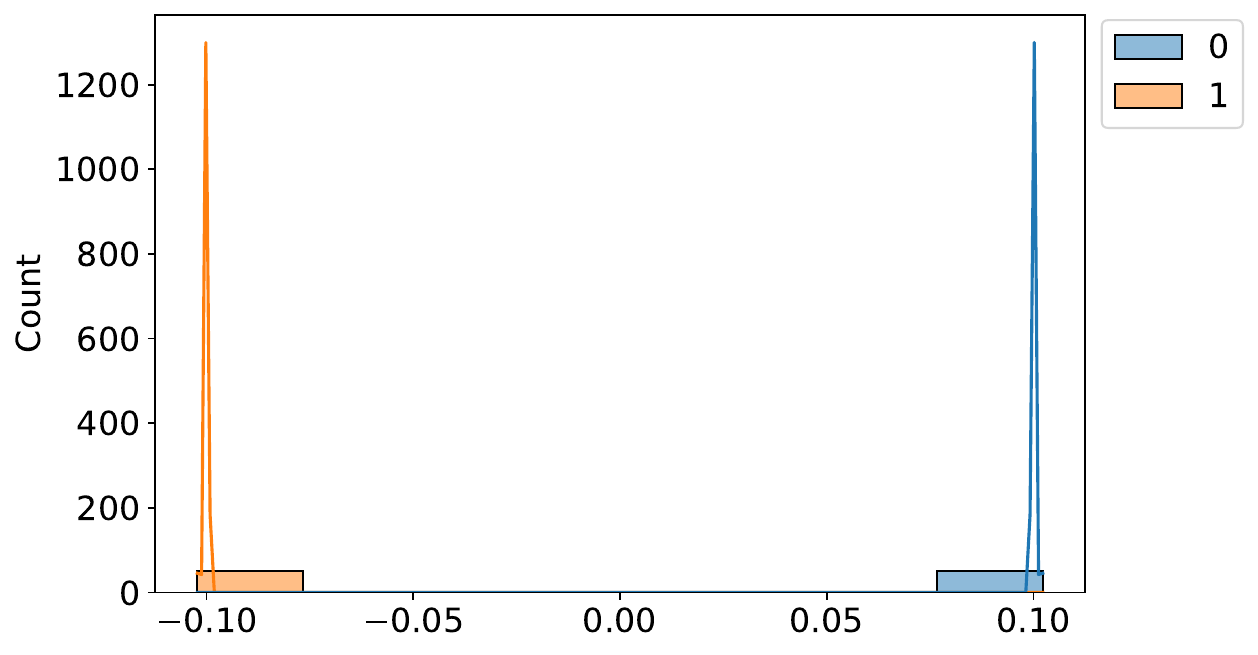}
         \caption{FDM scores histogram.}
     \end{subfigure}
    \caption{Top: FPCA and FDM scores in the first two components for the Moons dataset. Down: FPCA and FDM scores histogram  in the first component for Moons dataset.}
    \label{fig:moon-embedding}
\end{figure}

\begin{figure}
     \centering
     \begin{subfigure}[b]{0.45\textwidth}
         \centering
         \includegraphics[width=\textwidth,height=1.7in]{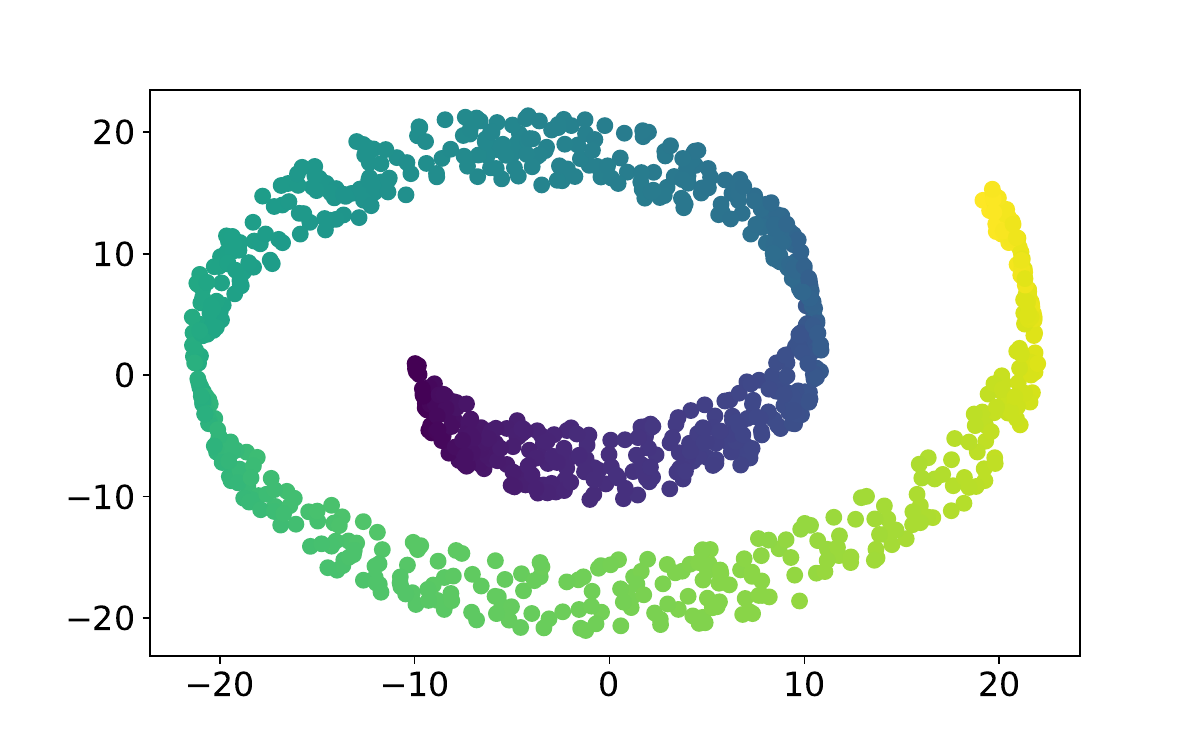}
         \caption{FPCA scores in 2D.}
     \end{subfigure}
     \hfill
     \begin{subfigure}[b]{0.45\textwidth}
         \centering
         \includegraphics[width=\textwidth,height=1.7in]{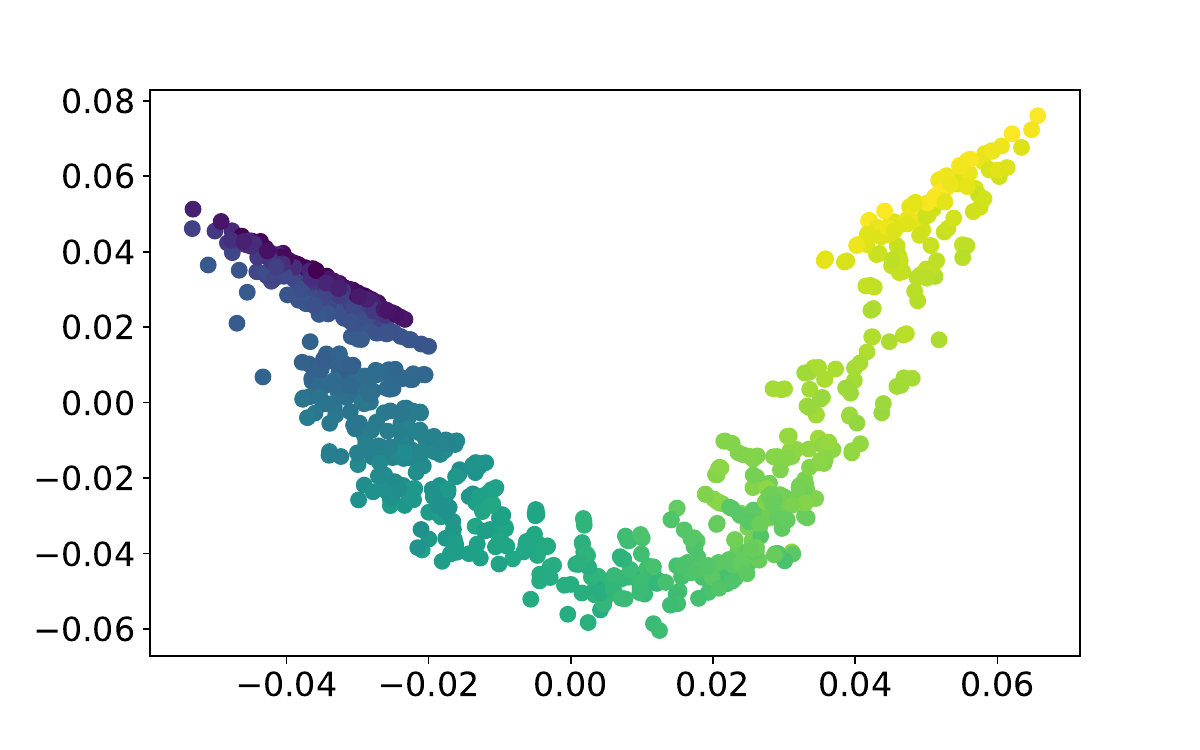}
         \caption{FDM scores in 2D.}
     \end{subfigure}
        \hfill
        \begin{subfigure}[b]{0.45\textwidth}
         \centering
         \includegraphics[width=\textwidth,height=1.7in]{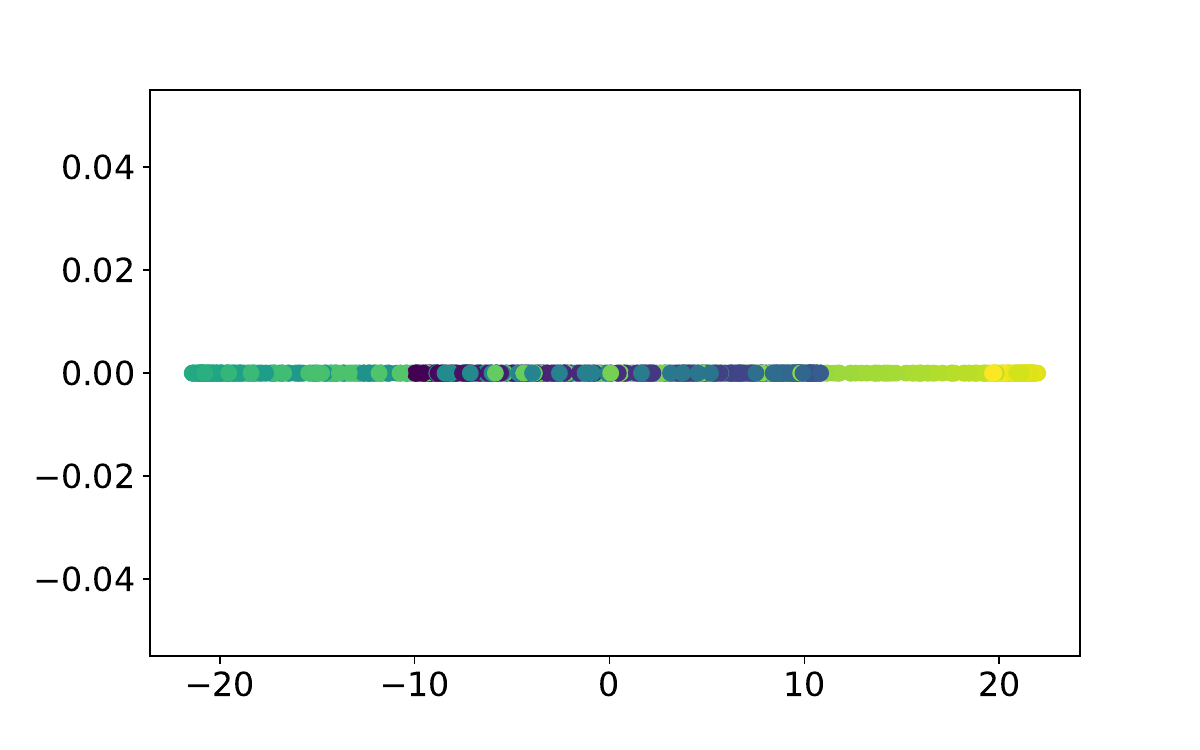}
         \caption{FPCA scores projection in 1D.}
     \end{subfigure}
     \hfill
     \begin{subfigure}[b]{0.45\textwidth}
         \centering
         \includegraphics[width=\textwidth,height=1.7in]{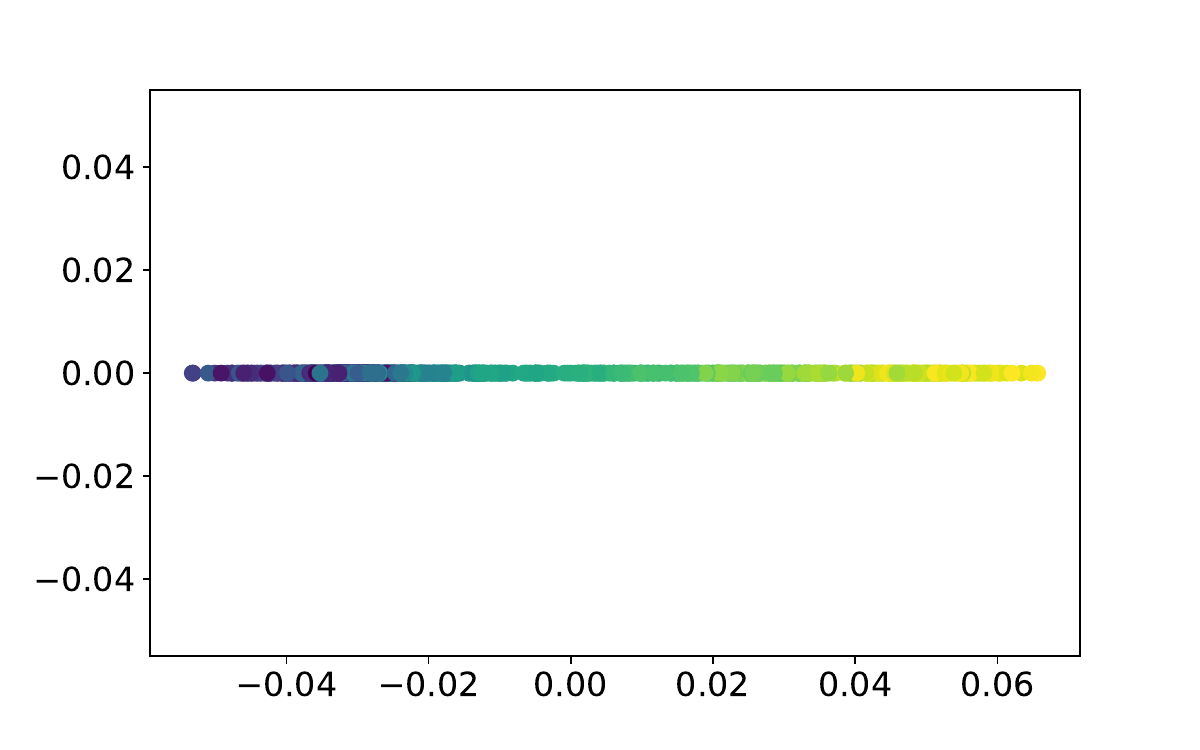}
         \caption{FDM scores projection in 1D.}
     \end{subfigure}
    \caption{FPCA and FDM scores for the Swiss Roll dataset.}
    \label{fig:swiss-roll-embedding}
\end{figure}

At the top panel of Figures~\ref{fig:moon-embedding} and~\ref{fig:swiss-roll-embedding}, the Moons and Swiss Roll scatterplots for the FPCA and FDM scores in the first two components are respectively depicted. 
Furthermore, below the same figures, the projection of the scores in the first component is also presented.
These visualizations enable us to discern whether clusters have been recognized or if the underlying manifold has been ``unrolled".

The FPCA embeddings exhibit similarity to the original Moons and Swiss Roll data, with the exception of rotation, sign, and scale adjustments. 
Instead, the FDM embeddings reveal that Moons data is entirely separated by the first component. Regarding the Swiss Roll, data, the underlying two-dimensional manifold is exposed, and even in one dimension it can be seen that color order is preserved by FDM. 

\subsection{Phoneme data}

In this experiment we have applied the main functional dimensionality reduction techniques---FPCA, Isomap and FDM---to a real dataset, namely the Phoneme dataset, which consists of log-periodograms computed from continuous speech of male speakers for five distinct phonemes.
This dataset was extracted from the TIMIT database~\cite{timit}, which is widely used in speech recognition research.
The phonemes are transcribed as follows: /sh/ (as in `she'), /dcl/ (as in `dark'), /iy/ (as the vowel in `she'), /aa/ (as the vowel in `dark') and /ao/ (as the first vowel in `water').
For our experiment, we take a random sample of 1500 log-periodograms curves of length 256 as a representative sample of the population due to computational cost.
Table~\ref{tab:phoeneme_info} shows the phoneme frequencies by class membership.

\begin{table}[ht]
    \centering
    \caption{Phoneme frequencies.}\label{tab:phoeneme_info}
    \begin{tabular}[t]{ccccc}
    \toprule
    \textbf{aa} & \textbf{ao} & \textbf{dcl} & \textbf{iy} & \textbf{sh} \\
    \toprule
    232 & 358 & 234 & 387 & 289 \\
    \bottomrule
    \end{tabular}
\end{table}

The log-periodogram curves of the Phoneme dataset are characterized by high irregularity and noise at their endpoints.
Hence, in order to prepare them for a dimensionality reduction analysis, a preprocessing step of trimming and smoothing is typically performed. 
To do this, we truncate the log-periodogram curves to a maximum length of 50.

\begin{figure}[ht]
    \centering
    \includegraphics[scale=0.33]{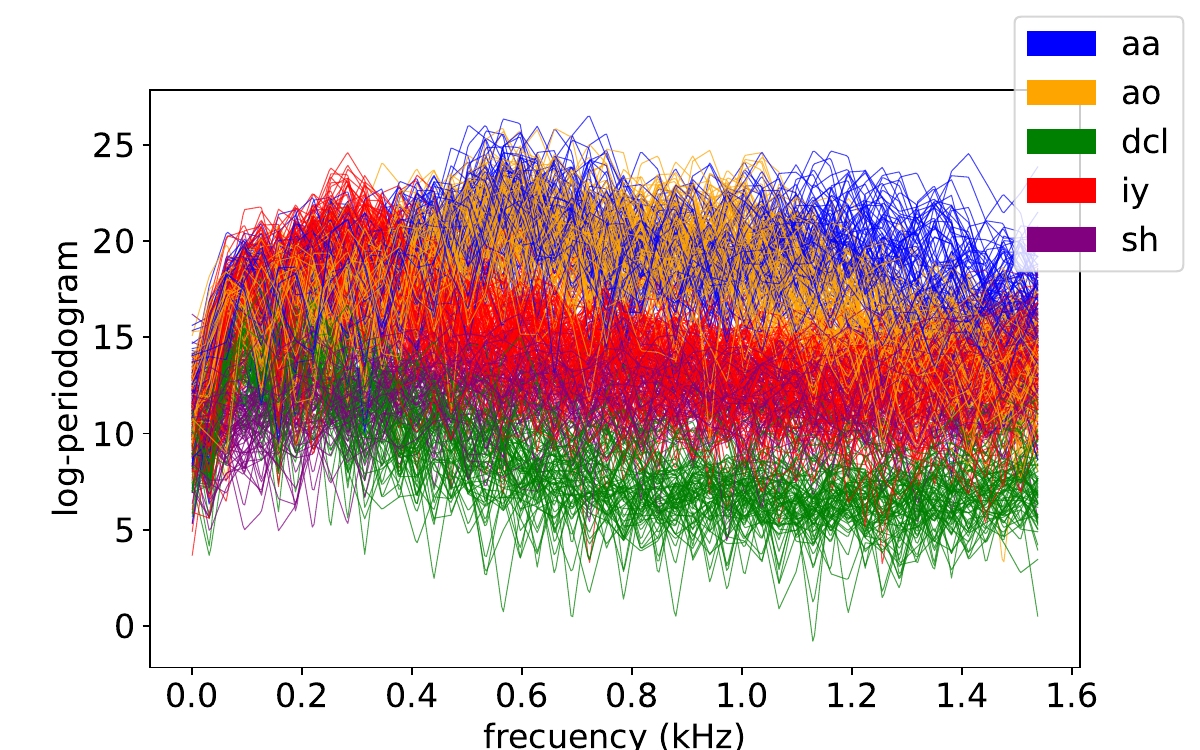}
    \caption{Phoneme dataset.}
    \label{fig:phoneme-fd-basis}
\end{figure}

Figure~\ref{fig:phoneme-fd-basis} displays the resulting functional representation of the phoneme dataset.

Looking at the curves, we see higher similarities between the phonemes /aa/ and /ao/, as their sound is very similar; and also between the last part of the curves of phonemes /iy/ and /sh/, since the phoneme /iy/ is contained at the end of the phoneme /sh/.
On the other hand, the curves of the phoneme /dcl/ seem to be far away from the rest of the curves.

Next, FPCA, Isomap and FDM are applied to the preprocessed dataset. 
In this experiment, both Isomap and FDM techniques require an initial analysis to determine the most suitable parameters.
After evaluating all possible parameter configurations, we found that a RBF kernel with $\sigma = 1.0$ and $\alpha = 1.0$ yields the best results for the FDM method, while for the Isomap method, we found that using 15 neighbors provides suitable performance.

\begin{figure}[ht]
     \centering
     \begin{subfigure}[b]{0.45\textwidth}
         \centering
         \includegraphics[width=\textwidth,height=1.3in]{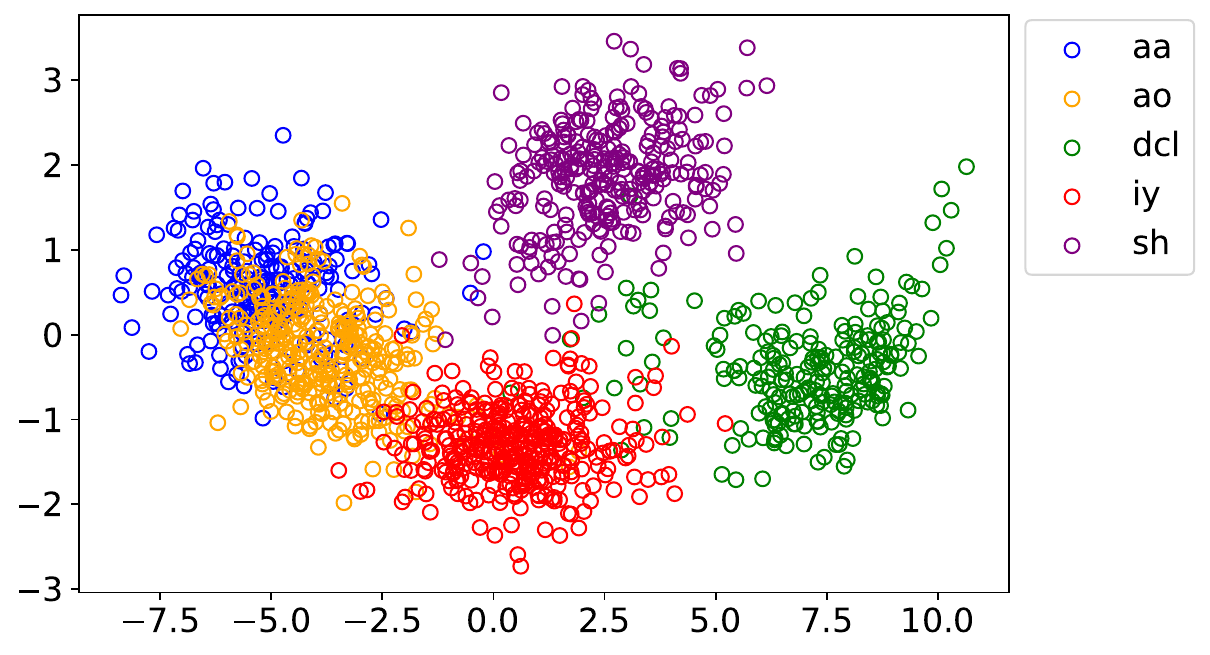}
         \caption{FPCA scores.}
     \end{subfigure}
     \begin{subfigure}[b]{0.45\textwidth}
         \centering
         \includegraphics[width=\textwidth,height=1.3in]{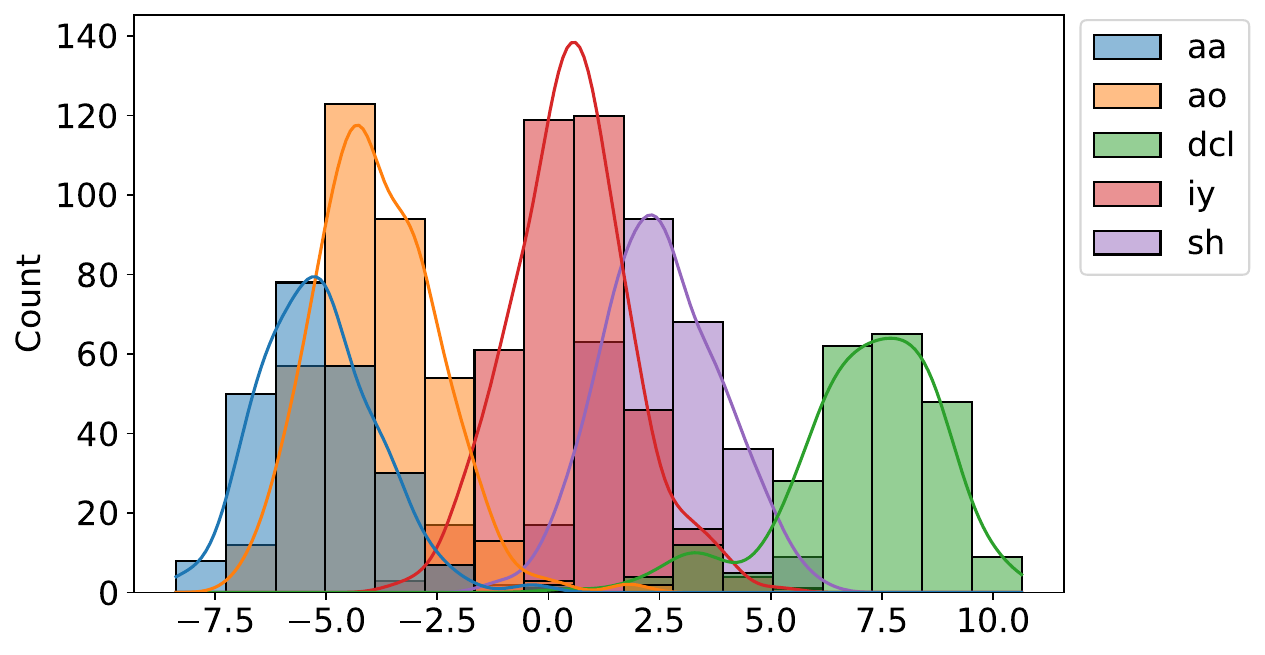}
         \caption{FPCA scores histogram.}
     \end{subfigure}
     \hfill
     \begin{subfigure}[b]{0.45\textwidth}
         \centering
         \includegraphics[width=\textwidth,height=1.3in]{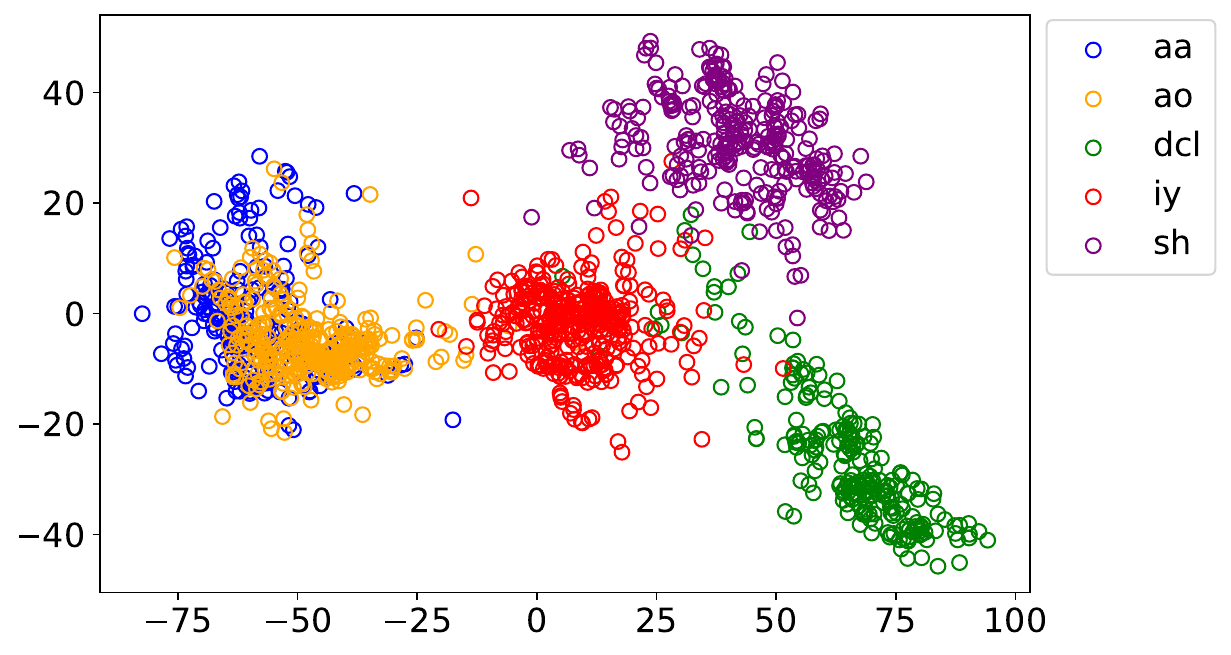}
         \caption{Isomap scores.}
     \end{subfigure}
     \begin{subfigure}[b]{0.45\textwidth}
         \centering
         \includegraphics[width=\textwidth,height=1.3in]{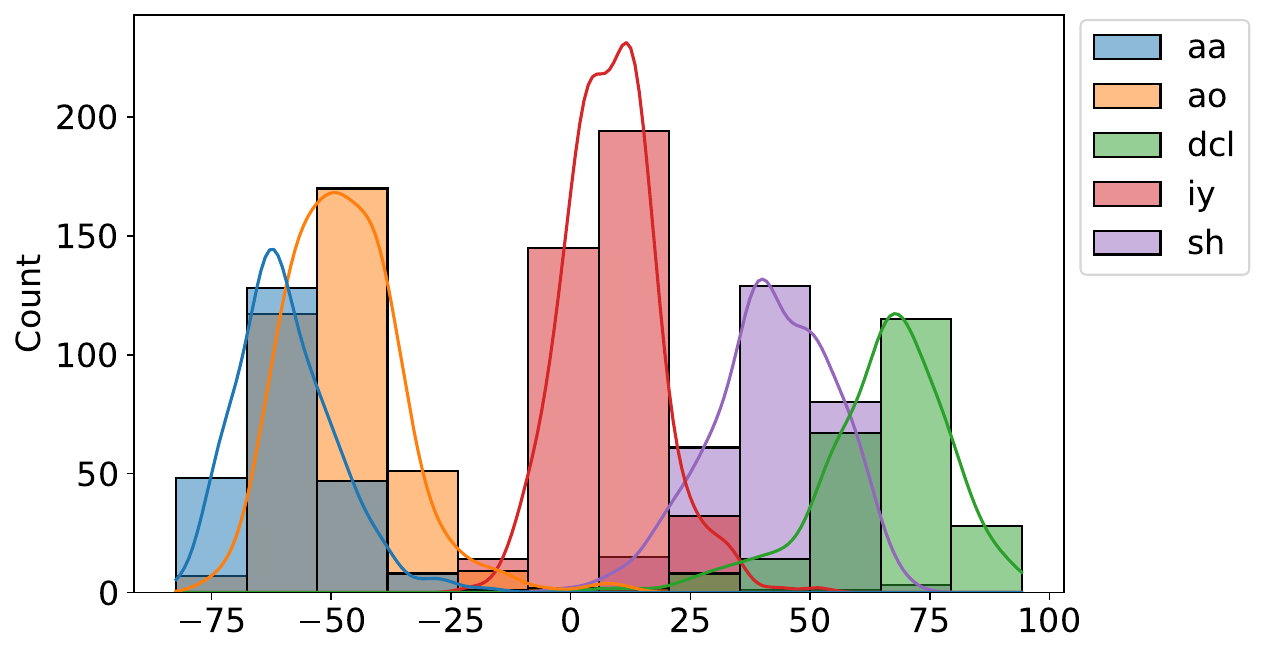}
         \caption{Isomap scores histogram.}
     \end{subfigure}
     \hfill
     \begin{subfigure}[b]{0.45\textwidth}
         \centering
         \includegraphics[width=\textwidth,height=1.3in]{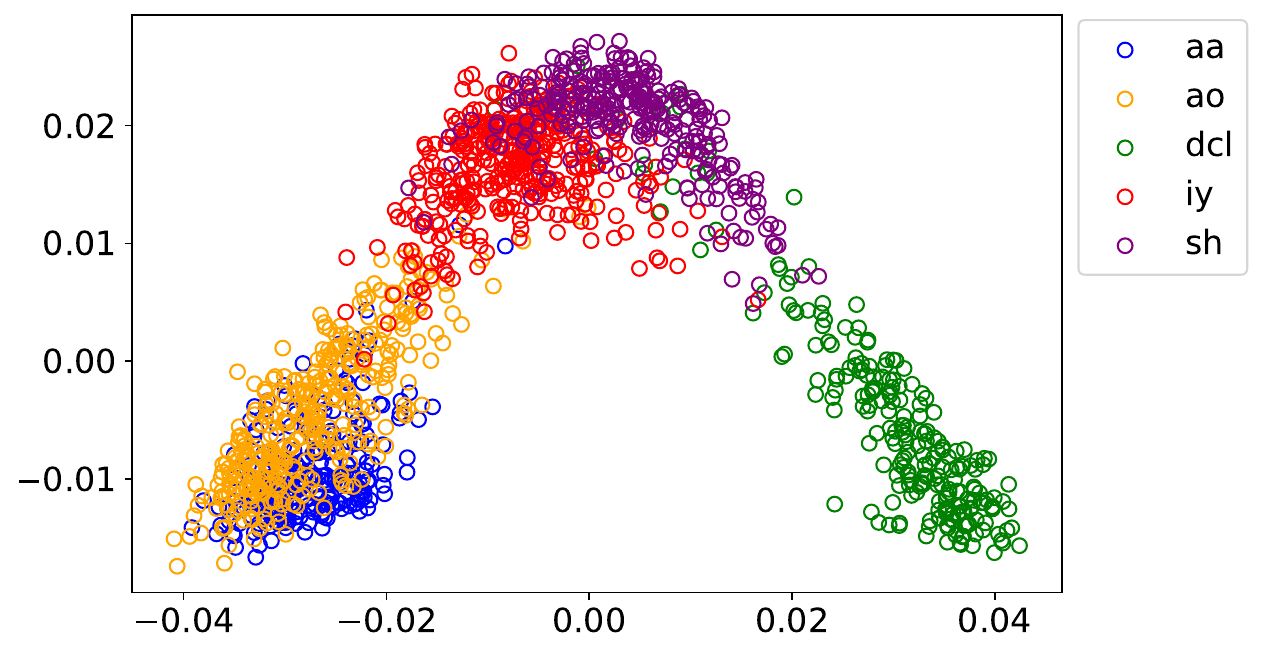}
         \caption{FDM scores.}
     \end{subfigure}
     \begin{subfigure}[b]{0.45\textwidth}
         \centering
         \includegraphics[width=\textwidth,height=1.3in]{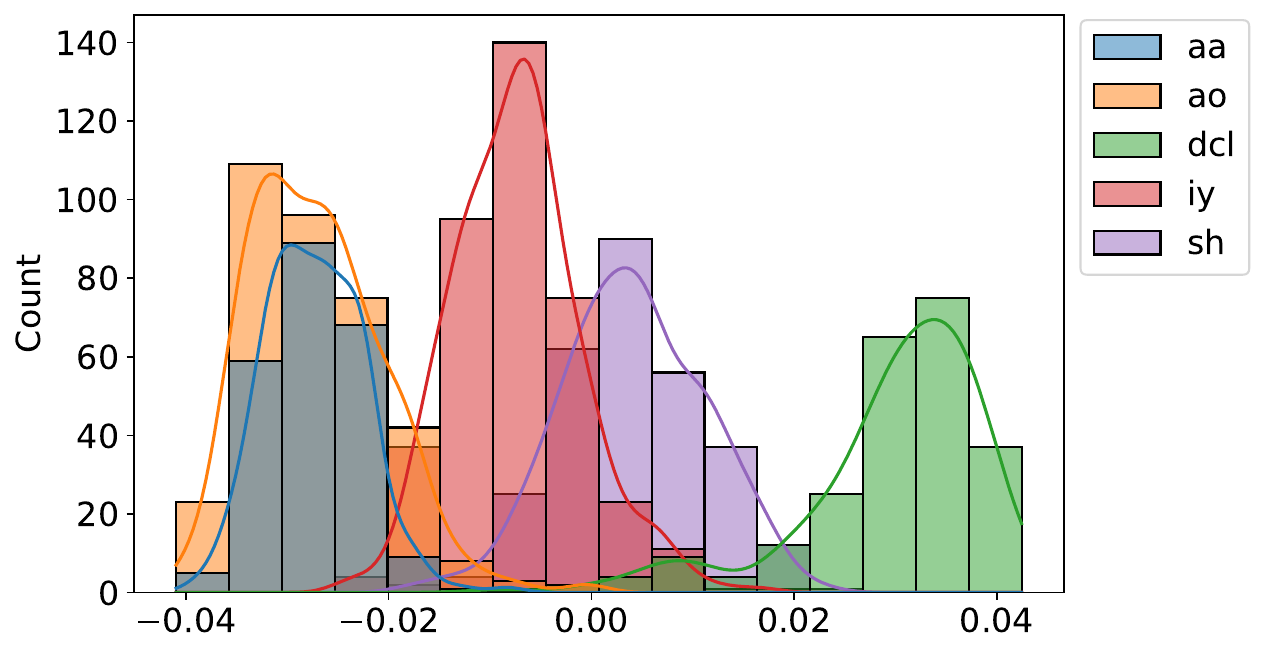}
         \caption{FDM scores histogram.}
     \end{subfigure}
     
        \caption{Left: FPCA, Isomap and FDM scores in the first two components for the Phoneme dataset. Right: FPCA, Isomap and FDM scores histogram  in the first component for Phoneme dataset.}
    \label{fig:phoneme-embedding}
\end{figure}

In the left panel of Figure~\ref{fig:phoneme-embedding}, we display the FPCA, Isomap and FDM scores in the first two components for the Phoneme dataset.
In the right panel of the same figure, we also present the histograms of the scores associated with the first component. The outcomes yielded by each technique offer different perspectives on phoneme similarities.

FPCA groups each phoneme maintaining the similarities that we had observed from the trajectories.
We can also observe in FPCA histogram the large overlapping between the first component of the phonemes, especially for /aa/ with /ao/ and /iy/ with /sh/ phonemes.

Isomap keep similar clusters than FPCA, but without overlapping /iy/ and /sh/ phonemes. 
Although a more effective separation of /iy/ and /sh/ is observed in the 2D embedding, there is a significant overlap between /dcl/ and /sh/ in the 1D embedding, despite the fact that they should be completely distinct.

FDM embeds the different phoneme groups ordering them from the vowel phonemes (/aa/, /ao/, /iy/) to the word phonemes (/sh/, /dcl/), in the same order as described. 
In addition, the similarity of the phonemes of the vowel of `dark' (/aa/) and the first vowel of `water' (/ao/) is maintained, appearing close in the embedding, as well as the similarity between the phoneme (/iy/) of the vowel of `she' and the phoneme (/sh/) corresponding to that word.
We can also observe that the phoneme of the vowel of `she' (/iy/) is more similar to the phoneme of the first vowel of `water' (/ao/) than to the phoneme of the vowel of `dark', information that we can observe in the curvature of both trajectories. 
This similarities may be because the phoneme /iy/ is more open than phonemes /ao/ and /iy/ and neither FPCA nor Isomap were able to capture this information.

\section{Conclusions}\label{sec:conclusion}

Functional dimensionality reduction methods are statistical techniques that represent infinite-dimensional data in lower dimensional spaces, for example, by capturing most of the variance of the original data in the new space or by reflecting lower dimensional underlying manifolds where the original data lay.
Following this second line of research, Diffusion Maps has been extended to the infinite-dimensional case.

As a result, Functional Diffusion Maps emerges as a functional manifold learning technique that finds low-dimensional representations of $L^2$ functions on non-linear functional manifolds.
FDM performs the same operations as DM algorithm once a functional similarity graph is created.
Even though giving a geometric interpretation of similarity between functional data is sometimes not possible, FDM retains the advantages of multivariate DM as it is robust to noise perturbation and it is also a very flexible algorithm that allows fine-tuning of parameters that influence the resulting embedding.

The performance of this method has been tested for simulated and real functional examples and the results have been compared with those obtained from the multivariate DM, the linear FPCA method, and the non-linear Isomap technique. 
It should be noted that the Isomap and multivariate DM cannot be applied directly to non-equispaced functions, as it may not effectively differentiate the particularities of the functions in study.
FDM outperforms FPCA for functions laying in non-linear manifolds such as the Moons and the Swiss Roll examples, where FDM obtains a representation that maintains the original structure of the manifold.
Besides being an advantage for these simulated examples, it also provides good results and allows new similarity interpretations for real examples such as the Phoneme dataset.

Overall, we find that the proposed manifold FDM method is an interesting technique that can provide useful representations which are competitive with, and often superior to, some classical linear and non-linear representations for functional data.

Nevertheless, some work remains to be done. In particular, the distance between functions can be interpreted as an earth mover distance between trajectories using a Besov space distance, which can be computed by expanding each function in a Haar basis in time, and minimizing its dual to Holder~\cite{ANKENMAN2018551}.

\backmatter

\bibliography{main}% common bib file


\end{document}